\documentclass[letterpaper]{article}
\usepackage{uai2020}
\usepackage[margin=1in]{geometry}
\usepackage{hyperref}
\usepackage{url, graphicx, subfigure, caption, dsfont, siunitx}
\usepackage{color, soul, eucal, makecell, booktabs, multirow, adjustbox}
\usepackage{amsmath,amsfonts,bm}
\usepackage{natbib}
\usepackage{enumitem}

\DeclareMathOperator*{\expec}{\mathbb{E}}
\DeclareMathOperator*{\amin}{\rm argmin}
\DeclareMathOperator*{\amax}{\rm argmax}
\newcommand{\aoadd}[1]{\textcolor{black}{#1}}
\newcommand{\dpadd}[1]{\textcolor{black}{#1}}

\newcommand{\green}[1]{\textcolor{green}{#1}}

\usepackage{times}

\title{GAN-based Priors for Quantifying Uncertainty}
\author{
\bf{Dhruv V. Patel} \\
University of Southern California \\
Los Angeles, CA, USA \\
\texttt{dhruvvpa@usc.edu}
\And
\bf{Assad A. Oberai} \\
University of Southern California \\
Los Angeles, CA, USA \\
\texttt{aoberai@usc.edu}
}

\begin{document}

\maketitle

\begin{abstract}
Bayesian inference is used extensively to quantify the uncertainty in an inferred field given the measurement of a related field when the two are linked by a mathematical model. Despite its many applications, Bayesian inference faces challenges when inferring fields that have discrete representations of large dimension, and/or have prior distributions that are difficult to characterize mathematically. In this work we demonstrate how the approximate distribution learned by a deep generative adversarial network (GAN) may be used as a prior in a Bayesian update to address both these challenges.  We demonstrate the efficacy of this approach on two distinct, and remarkably broad, classes of problems. The first class leads to supervised learning algorithms for  image classification with superior out of distribution detection and accuracy, and for image inpainting with built-in variance estimation. The second class leads to unsupervised learning algorithms for  image denoising and for  solving physics-driven inverse problems.

\end{abstract}

\section{INTRODUCTION}

Quantifying uncertainty in an inference problem amounts to making a prediction and quantifying the confidence in that prediction. In the context of an image recovery problem, this may be understood as follows. A typical computer vision algorithm uses a noisy version of an image and prior knowledge to produce the  inferred image which can be interpreted as the ``best guess'' of the original image. Quantifying uncertainty in this context involves generating an estimate of the level of confidence in the best guess, in addition to the guess itself. 


Bayesian inference provides a principled approach for quantifying uncertainty. As shown in the following section, it treats the inferred vector as a multivariate stochastic vector and leads to an expression for its distribution. This expression can be used to estimate the most likely solution (the maximum a-posteriori estimate, or the MAP), the mean, the variance, or any other population parameter of interest. Thus  \dpadd{providing} a recipe for thoroughly quantifying the uncertainty in an inference problem. For the image recovery problems considered in this paper, Bayesian inference not only provides the best guess of the true image, but also a means to estimate measures of uncertainty such as the pixel-wise variance or the likelihood of an out-of-distribution input. 

The knowledge of uncertainty in a prediction can directly influence the downstream action that depends on the inference. Consider an image recovery problem where two distinct inputs lead to similar recovered images: those of a traffic sign with a high speed limit. However, for the first input the predicted variance is small, while for the second input it is large. Further, the set of likely images in the second set also includes images of a Stop Sign. Then the appropriate action for the two inputs, determined after solving the inference problem and quantifying uncertainty, is very different. For the first input, the appropriate action is one of continued motion, whereas for the second input it is to slow down. Similar examples can be drawn from other areas like medical imaging, high frequency trading and autonomous systems (\cite{gal2016uncertainty}).

\begin{figure*}[!ht]
\begin{center}
\includegraphics[width=\linewidth]{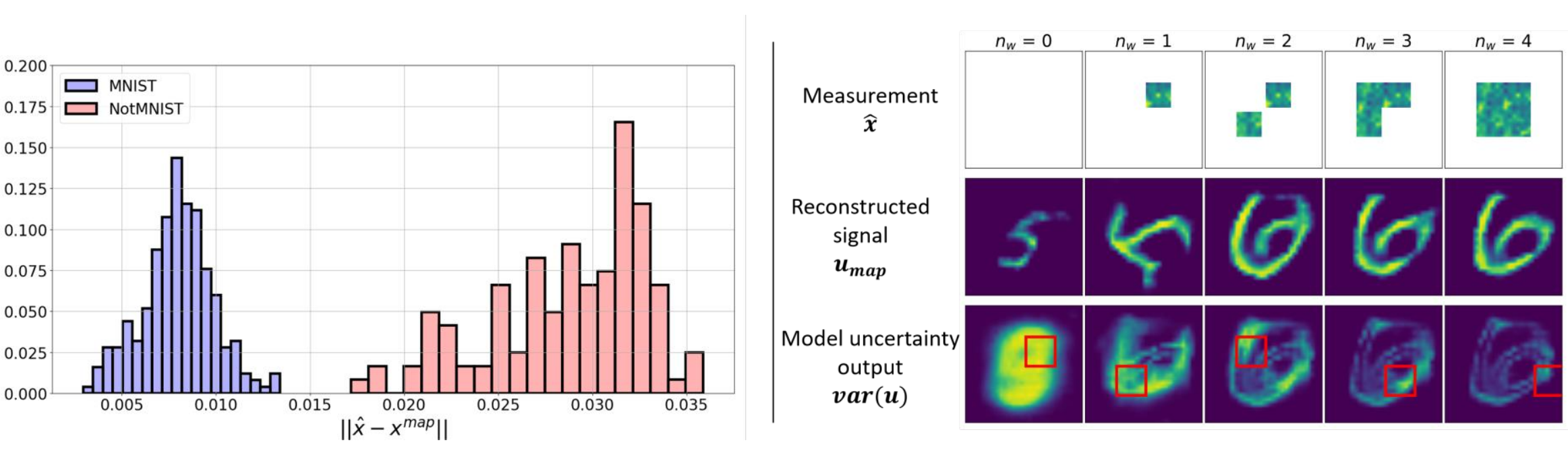} 
\end{center}
\caption{Left panel: Histogram of $ \| \hat{\bm{x}} - \bm{x}^{\rm map}\|$, our measure for out-of-distribution (OOD) data detection, on classification experiments on MNIST. The proposed method is able to successfully distinguish in-distribution (MNIST) and OOD (NotMNIST) test inputs. A large value of this parameter is a warning  to the end user to disregard the classification results. Right panel: Estimate of the MAP (2nd row) and pixel-wise variance (3rd row) from the limited view of a noisy image (1st row) using the proposed method for image inpainting with a prior trained on MNIST images. An active learning strategy based on the maximum value of variance is used to determine the location of the subsequent window. An accurate reconstruction of the original image is obtained with just 4 windows.} 
\label{fig:intro} 
\end{figure*}

The knowledge of uncertainty can also be \dpadd{useful in determining the optimal location of a sensor. Consider an image recovery problem, where the goal is to infer the signal, and associated uncertainty, using limited amount of measurement data. In this problem a user can leverage information about the spatial distribution of uncertainty to choose the location with maximum uncertainty as next measurement location.}
This \dpadd{task} falls within the fields of active learning and\dpadd{/or} design of experiments (\cite{degroot1962uncertainty}) and is particularly useful in applications like satellite imaging, where each measurement requires significant time and/or resources. 


In Figure \ref{fig:intro}, we demonstrate how the proposed GAN-based Bayesian inference algorithm for quantifying uncertainty is useful in the scenarios described above. In the first example it is used to compute a measure that detects, with perfect accuracy, that the input to an image classification algorithm is not from the dataset that was used to train it - the so-called out of distribution (OOD) detection problem. In the second example it computes the pixel-wise variance in an image inpainting task, which is then used to determine the location of the subsequent window to be revealed in an optimal iterative strategy.  We return to these applications in greater detail in Section 4.


\subsection{BAYESIAN INFERENCE} 

Bayesian inference is a well-established technique for quantifying uncertainties in inference problems (\cite{dashti2016bayesian}, \cite{kaipio2006statistical}). It has found applications in diverse fields  such as geophysics,
climate modeling,
chemical kinetics,
heat conduction,
astrophysics,
materials modeling,
and the detection and diagnosis of disease.
The two critical ingredients of this technique are - an informative prior distribution representing the prior belief about the parameters to be inferred and an efficient method for sampling from the posterior distribution. In this manuscript we describe how deep generative adversarial networks (GANs) can be effectively used in these roles.

We consider the setting where we wish to infer a vector of parameters $\bm{y} \in \mathbb{R}^N$ from the measurement of a related vector $\bm{x} \in \mathbb{R}^P$. We allow for two broad classes of problems. In one class (labeled Class 1 in Section 2) the map from $\bm{y}$ to $\bm{x}$ is known through a forward model $\bm{x} = \bm{f}(\bm{y})$. These types of problems are often referred to as inverse problems. In the other class (labeled  Class 2 in Section 2) this map is not known and must be inferred from prior data. In the discussion that follows, we apply Bayesian inference to Class 1 problems and point out two main challenges. We note that the same challenges apply to problems in Class 2 as well. 

 A noisy measurement of $\bm{x}$ is denoted by $\hat{\bm{x}} = \bm{f}(\bm{y}) + \bm{\eta}$, where the vector $\bm{\eta}  \in \mathbb{R}^P$ represents noise.  While the forward map $\bm{f}$ is typically well-posed, its inverse is not, and hence to infer $\bm{y}$ from the measurement $\hat{\bm{x}}$ requires techniques that account for this ill-posedness. Classical techniques based on regularization tackle this ill-posedness by using additional information about the sought solution field explicitly or implicitly (\cite{tarantola2005inverse}). Bayesian inference offers a different approach to this problem by modeling the unknown solution as well as measurements as random variables. This addresses the ill-posedness of the inverse problem, and allows for the characterization of the uncertainty in the inferred solution.

The notion of a prior distribution plays a key role in Bayesian inference.  Through multiple observations of the field $\bm{y}$, denoted by the set $\mathcal{S} = \{\bm{y}^{(1)}, \cdots, \bm{y}^{(S)}\}$, we have some prior knowledge of $\bm{y}$ that can be utilized when inferring $\bm{y}$ from $\hat{\bm{x}}$. This is used to build, or intuit, a prior distribution for $\bm{y}$, denoted by $p^{\rm prior}_{Y}(\bm{y})$. Some typical examples include Gaussian process prior with specified co-variance kernels, Gaussian Markov random fields, 
Gaussian priors defined through differential operators, 
and hierarchical Gaussian priors.
These priors promote smoothness \dpadd{and/}or structure in the inferred solution and can be expressed explicitly in an analytical form.

Another key component of Bayesian inference is a distribution that represents the likelihood of $\bm{x}$ given an instance of $\bm{y}$, denoted by $p^{\rm l}(\bm{x}|\bm{y})$. This is often determined by the distribution of the error in the model.
Given this, 
the posterior distribution of $\bm{y}$, determined using Bayes' rule after accounting for the observation $\hat{\bm{x}}$ is given by, 
\begin{eqnarray}
p^{\rm post}_{Y}(\bm{y}|\bm{x}) &=& \frac{1}{\mathbb{Z}} p^{\rm l}(\bm{x}|\bm{y}) p^{\rm prior}_{Y}(\bm{y}) 
\label{eq:bayes}
\end{eqnarray}
Here, $\mathbb{Z}$
is the prior-predictive distribution of $\bm{y}$.

The posterior distribution characterizes the uncertainty in $\bm{y}$; however for vectors of large dimension characterizing this distribution explicitly is a challenging task. Consequently the expression above is used to perform tasks that are more manageable. These include determining estimates such as the maximum a-posteriori estimate (MAP), expanding the posterior distribution in terms of other distributions that are simpler to work with (\cite{Bui-Thanh2012}), or using techniques like Markov Chain Monte-Carlo (MCMC) to generate samples that are close to the samples generated by the true posterior distribution (\cite{parno2018transport}).

In summary, despite its numerous applications, Bayesian inference faces significant challenges. These include defining a reliable and informative prior distribution for $\bm{x}$ when the set $\mathcal{S}$  is difficult to characterize \dpadd{analytically}, and efficiently sampling from the posterior distribution when the dimension of $\bm{x}$ is large; a typical situation in many practical science and engineering applications.

\subsection{OUR CONTRIBUTION}
\dpadd{The main contribution of this paper are}:
\begin{enumerate}[wide, labelwidth=!, labelindent=0pt]

    \item A novel method for performing Bayesian inference involving complex priors and high dimensional posterior. We utilize the distribution learned by a GAN as a surrogate for the prior distribution and reformulate the inference problem in the low-dimensional latent space of the GAN. Furthermore, we provide a theoretical analysis of the weak convergence of the posterior density learned by the proposed method to the true posterior density.
    \item Application of this method to problems where the map from the inferred to the measured vector is known {\em a-priori}. This leads to novel unsupervised algorithms for physics-based inverse problems and image denoising problems with quantitative measures of uncertainty. 
    \item Application of this method to problems where the map from the inferred to the measured vector is not known and is determined from data. This leads to novel algorithms for image classification and image inpainting with quantitative measures of uncertainty. 
    \item Demonstration of the utility of quantifying uncertainty in the detection of out-of-distribution (OOD) samples and in active learning. 
\end{enumerate}

\subsection{RELATED WORK}

The main idea developed in this paper tackles the challenges described above by training a generative adversarial network (GAN) using the sample set $\mathcal{S}$, and then using the distribution learned by the GAN as the prior distribution in Bayesian inference. 
Related work in this area can be organized by considering the two broad classes of problems this idea is applied to. 

In one class of problems, which are referred to as inverse problems, the map $\bm{x} = \bm{f}(\bm{y})$, that is the map from the inferred field to the measurement is known. The use of sample-based priors for solving inverse problems has a rich history (\cite{Calvetti2005}). As does the idea of reducing the dimension of the parameter space by mapping it to a lower-dimensional space (\cite{Marzouk2009}). However, the use of learning-based deep generative models like GANs in these tasks is novel. Recently, several authors have considered the use of learning-based methods for solving inverse problems arising in different domains. These include the use of deep convolutional neural networks (CNNs) or recurrent neural networks (RNNs) to solve physics-driven inverse problems (\cite{Adler2017, patel2019circumventing, Pesah2018}) and use of deep generative models like VAEs and GANs to solve inverse problems arising in computer vision (\cite{Kupyn2018a, CycleGAN2017, Chang}). There is also a growing body of work dedicated to using GANs as a regularizer in solving inverse problems (\cite{lunz2018adversarial} and in compressed sensing (\cite{bora2018ambientgan, kabkab2018task, shah2018solving}). However, these approaches differ from ours in that they solve the inverse problem as an optimization problem and do not rely on Bayesian inference; as a result, they add regularization in an ad-hoc manner and do not attempt to quantify the uncertainty in the inferred field. 
More recently, the approach described in (\cite{adler2018deep}) utilizes GANs in a Bayesian setting; however the GAN is trained to approximate the posterior distribution and not the prior, as in our case.

In another class of problems the forward map is not known; however in its lieu pair-wise instances of $\bm{y}$ and $\bm{x}$ are available. In this case the algorithms most closely related to our approach are the so-called hybrid methods, where an invertible  generator is trained to learn $p(\bm{x})$ and is linked to another network that is trained to produce $p(\bm{y}|\bm{x})$ (\cite{Nalisnick2019,Chen2019}). This algorithm is then applied to image classification problems, where for a given input image $\bm{x}$, $p(\bm{x})$ is used to determine the likelihood of the input and $p(\bm{y}|\bm{x})$  is used to infer the probability of the corresponding label. In contrast to this, we train a Wasserstein GAN to learn the joint density $p(\bm{u})$, where $\bm{u} = [\bm{x},\bm{y}]$ and then use this as the prior in a Bayesian update of the posterior density for a given input $\bm{x}$. The sampling problem for the posterior is reduced to the latent space of the GAN, which is of smaller dimension, and a Markov Chain Monte-Carlo algorithm is trained to generate samples of  $p(\bm{u}|\bm{x})$ and fully characterize the posterior density. 

We note that deep learning based Bayesian networks, where the network weights are stochastic parameters that are determined using Bayesian inference, are another means of quantifying uncertainty (\cite{gal2016dropout}), \aoadd{and have recently been applied to semantic image-segmentation} \dpadd{and super-resolution (\cite{ Kendall2019})}. 

\section{PROBLEM FORMULATION}
\label{S:2}

We consider the problem where we wish to infer the vector $\bm{y}$ from the noisy measurement of a related vector $\bm{x}$.
We consider two broad classes of problems of this type. In one class we assume that the forward operator which maps $\bm{y}$ to $\bm{x}$, that is $ \bm{x} = \bm{f}(\bm{y})$, is known. While in the other, we assume that $\bm{f}(\bm{y})$ is not known and any relation between $\bm{x}$ and $\bm{y}$ must be determined from data.

\subsection{CLASS 1: THE MAP $f(y)$ IS KNOWN}
These problems are commonly referred to as inverse problems and  the map $\bm{x} = \bm{f}(\bm{y})$ is called the forward map. In this class of problems, this forward map is usually well-defined and is assumed to be known either through physics-based principles or through other modeling paradigms.

A noisy measurement of $\bm{x}$ is denoted by $\hat{\bm{x}} = \bm{f}(\bm{y}) + \bm{\eta}$, where $\bm{\eta} \sim p_\eta$ is the noise vector. In addition, it is assumed that the sample set $\mathcal{S} = \{\bm{y}^{(1)}, \cdots, \bm{y}^{(S)}\} $, which contains multiple  realizations of $\bm{y}$ drawn from the distribution $P_Y$, is also known. The goal is to use the prior information encoded in $\mathcal{S}$, the noisy measurement $\hat{\bm{x}}$, and the forward map $\bm{f}$ to determine the distribution of the vector $\bm{y}$.

The prior information for this class of problems is built from the distribution of $\bm{y}$ alone. Thus problems in this class fall into unsupervised learning category,  where training only requires instances of one type of data (the vector to be inferred). The need to have access to pairwise samples of $\bm{x}$ and $\bm{y}$ is circumvented by the knowledge of the forward operator. An example problem in this class is that of image denoising, where $\bm{x}$ represents a noisy image, the operator $\bm{f}$ is the identity, and $\bm{y}$ is the de-noised image. Another example, which is drawn from physics-based inverse problems, is where one wishes to infer the initial temperature field from a measurement of the temperature field at a later time. Here $\bm{x}$ represents the temperature field at a finite time $T > 0$, $\bm{f}$ is the forward in time heat conduction operator, and $\bm{y}$ is the temperature at the initial time. A large number of other physics-based inverse problems can also be cast in this form.

Given $\hat{\bm{x}}$, using Bayes' rule we may write the posterior distribution of $\bm{y}$ as,
\begin{eqnarray}
p^{\rm post}_{Y}(\bm{y}|\bm{x}) &=& \frac{1}{\mathbb{Z}} p^{\rm l}(\bm{x}|\bm{y}) p^{\rm prior}_{Y}(\bm{y}) \nonumber \\ &=& \frac{1}{\mathbb{Z}} p_\eta(\hat{\bm{x}} - \bm{f}(\bm{y})) p_{Y}(\bm{y}). \label{eq:bayesy}
\end{eqnarray}
where $\mathbb{Z}$ is the prior-predictive distribution of $\bm{y}$ and ensures that the posterior integrates to one.

In order to efficiently sample the posterior density, we first train a GAN using the set $\mathcal{S}$ whose elements are sampled from $P_Y$.  We let $\bm{z} \sim p_Z(\bm{z})$ characterize the latent vector space of the GAN, and let $\bm{g}(\bm{z})$ and $d(\bm{y})$ denote its generator and discriminator, respectively. In Appendix \ref{sec:appweak}, assuming that (a) the stationarity conditions for the adversarial loss function are satisfied and (b) that the set of basis functions obtained by taking the derivative of the discriminator with respect to its weights forms a complete basis in $L^\infty(\Omega_y)$, in the limit of infinite capacity, we prove that  
\begin{eqnarray}
\expec_{\bm{y} \sim p_Y} [m(\bm{y})] = \expec_{\bm{z} \sim p_Z} [ m (\bm{g}(\bm{z}))], \label{eq:emy}
\end{eqnarray}
for sufficiently smooth $m(\bm{y})$. This equation states that once a GAN is trained using the sample set $\mathcal{S}$, it may be used to evaluate any population parameter for $p_Y$ by sampling from $p_Z$ and then passing the samples through the generator. Since the dimension of the latent space is much smaller than that of $\bm{y}$\green{,} this represents an efficient means of evaluating population parameters.

In order to turn this into an expression for evaluating  a population parameter for the posterior density, we select $m(\bm{y}) = \frac{l(\bm{y}) p_\eta(\hat{\bm{x}} - \bm{f}(\bm{y}))}{\mathbb{Z}}$, substitute this expression on both side of (\ref{eq:emy}), and use (\ref{eq:bayesy}) to arrive at,
\begin{eqnarray}
\expec_{\bm{y} \sim p^{\rm post}_{Y}}[l(\bm{y})] 
    &=&  \expec_{\bm{z} \sim p^{\rm post}_{Z}}[l(\bm{g}(\bm{z}))], \label{eq:pypost}
\end{eqnarray}
where 
\begin{eqnarray}
p^{\rm post}_Z(\bm{z}|\bm{x}) \equiv \frac{1}{\mathbb{Z}} p_\eta(\hat{\bm{x}} - \bm{f}(\bm{g}(\bm{z}))) p_{Z}(\bm{z}). \label{eq:pzposty}
\end{eqnarray}
The distribution $p^{\rm post}_Z$ is the analog of $p^{\rm post}_Y$ in the latent vector space. The measurement $\hat{\bm{x}}$ updates the prior distribution for $\bm{y}$ to the posterior distribution. Similarly, it updates the prior distribution for $\bm{z}$, $p_Z$, to the posterior distribution, $p^{\rm post}_Z$, defined above. 

Equation (\ref{eq:pypost}) implies that sampling from the posterior distribution of $\bm{y}$ is equivalent to sampling from the posterior distribution for $\bm{z}$ and passing the sample through the generator $\bm{g}$. That is,
\begin{eqnarray}
\bm{y} \sim p^{\rm post}_Y (\bm{y}| \bm{\hat{x}}) \Rightarrow \bm{y} = \bm{g}(\bm{z}), \bm{z} \sim p^{\rm post}_Z(\bm{z}|\bm{\hat{x}}). 
\end{eqnarray}
Since the dimension of $\bm{z}$ is typically smaller than that of $\bm{y}$, this represents an efficient approach to sampling from the posterior of $\bm{y}$. 

The left hand side of (\ref{eq:pypost}) is an expression for a population parameter of the posterior.
The right hand side of this equation describes how this parameter may be evaluated by sampling $\bm{z}$ (instead of $\bm{y}$) from  $p_Z^{\rm post}$. In practise this is accomplished by generating an MCMC approximation, $p^{\rm mcmc}_Z(\bm{z}|\bm{x}) \approx p^{\rm post}_Z(\bm{z}| \bm{x})$ using the definition in (\ref{eq:pzposty}), and thereafter sampling $\bm{z}$ from this distribution. This circumvents the calculation of the prior-predictive distribution of $\bm{y}$ (denoted by $\mathbb{Z})$, which would be necessary when using (\ref{eq:pzposty}) directly. Then from (\ref{eq:pypost}), any desired population parameter for posterior distribution may be approximated as 
\begin{eqnarray}
 \overline{l (\bm{y})} &\equiv& \expec_{\bm{y} \sim p^{\rm post}_{Y}}[l(\bm{y})]  \nonumber \\
 &\approx& \frac{\sum_{n = 1}^{N_{\rm samp}} l(\bm{g}(\bm{z}))}{N_{\rm samp}}, \quad \bm{z} \sim  p^{\rm mcmc}_Z(\bm{z}| \bm{x}).  \label{eq:mcmcy} 
\end{eqnarray}
For all the numerical experiments in this paper we have used this approach to evaluate population parameters.

\subsection{CLASS 2: THE MAP $f(y)$ IS NOT KNOWN}
We now consider problems where the relation between $\bm{x}$ and $\bm{y}$ is not known and must be inferred from data. 
We denote by $\bm{u} = [\bm{x},\bm{y}]$ the joint vector and recognize that the measurement has the form $\hat{\bm{x}} = \mathds{1}_{\bm{x}} \bm{u} + \bm{\eta}$, where $\mathds{1}_{\bm{x}}$ is the indicator function that extracts  components of $\bm{x}$ from $\bm{u}$, and $\bm{\eta}$ is  the noise vector drawn from the distribution $p_\eta$. Further we assume that the sample set $\mathcal{S}= \{\bm{u}^{(1)}, \cdots, \bm{u}^{(S)}\}$ contains multiple measurements of $\bm{u}$ drawn from the distribution $P_U$. The goal is to use the prior information encoded in $\mathcal{S}$ and the new, noisy measurement $\hat{\bm{x}}$ to determine the distribution for the corresponding vector $\bm{y}$, and perhaps also the de-noised version of $\bm{x}$.

Since the prior information is built from the joint distribution $P_U$ this class of problems is one of supervised learning where training requires \textit{pair-wise} instances of $\bm{x}$ and $\bm{y}$. An example problem in this class is that of image classification, where $\bm{x}$ represents an image and $\bm{y}$ represents the corresponding one-hot encoded label vector. Another example is that of image inpainting, where $\bm{x}$ represents the portion of an image that is revealed and $\bm{y}$ represents the portion that is occluded. 

The use of GANs as priors in this class of problems closely parallels the development for problems treated in the previous section. Therefore, rather than repeating the entire development below, we only highlight the important steps and salient differences. 

Using Bayes' rule we may write the posterior distribution of $\bm{u} = [\bm{x}, \bm{y}]$ as,
\begin{eqnarray}
p^{\rm post}_{U}(\bm{u}|\bm{x}) &=& \frac{1}{\mathbb{Z}} p^{\rm l}(\bm{x}|\bm{u}) p^{\rm prior}_{U}(\bm{u}) \nonumber \\ &=& \frac{1}{\mathbb{Z}} p_\eta(\hat{\bm{x}} - \mathds{1}_{\bm{x}} (\bm{u})) p_{U}(\bm{u}), \label{eq:bayesu}
\end{eqnarray}
where $\mathbb{Z}$
is the prior-predictive distribution of $\bm{u}$. In order to efficiently sample the posterior density we train a GAN using the set $\mathcal{S}$ to generate a prior.  As before, we let $\bm{z} \sim p_Z(\bm{z})$ characterize the latent vector space of the GAN, and let $\bm{g}(\bm{z})$ denote its generator. 
Under the assumptions of the result derived in  Appendix \ref{sec:appweak}, we have 
\begin{eqnarray}
\expec_{\bm{u} \sim p_U} [m(\bm{u})] = \expec_{\bm{z} \sim p_Z} [ m (\bm{g}(\bm{z}))], \label{eq:emu}
\end{eqnarray}
for sufficiently smooth $m(\bm{u})$. We choose $m(\bm{u}) = \frac{l(\bm{u}) p_\eta(\hat{\bm{x}} - \mathds{1}_{\bm{x}} (\bm{u}))}{\mathbb{Z}}$, substitute it in (\ref{eq:emu}), and make use of (\ref{eq:bayesu})  to arrive at,
\begin{eqnarray}
\expec_{\bm{u} \sim p^{\rm post}_{U}}[l(\bm{u})] 
    &=&  \expec_{\bm{z} \sim p^{\rm post}_{Z}}[l(\bm{g}(\bm{z}))], \label{eq:pupost}
\end{eqnarray}
where 
\begin{eqnarray}
p^{\rm post}_Z(\bm{z}|\bm{x}) \equiv \frac{1}{\mathbb{Z}} p_\eta(\hat{\bm{x}} - \mathds{1}_{\bm{x}} (\bm{g}(\bm{z}))) p_{Z}(\bm{z}). \label{eq:pzpostu}
\end{eqnarray}
The distribution $p^{\rm post}_Z$ is the analog of $p^{\rm post}_U$ in the latent vector space. We use (\ref{eq:pzpostu}) to generate an MCMC approximation, $p^{\rm mcmc}_Z(\bm{z}|\bm{x}) \approx p^{\rm post}_Z(\bm{z}| \bm{x})$ of the posterior. Thereafter, from (\ref{eq:pupost}) we conclude that any population parameter for the posterior can be approximated as 
\begin{eqnarray}
 \overline{l (\bm{u})} &\equiv& \expec_{\bm{u} \sim p^{\rm post}_{U}}[l(\bm{u})]  \nonumber \\
 &\approx& \frac{\sum_{n = 1}^{N_{\rm samp}} l(\bm{g}(\bm{z}))}{N_{\rm samp}}, \quad \bm{z} \sim  p^{\rm mcmc}_Z(\bm{z}| \bm{x}).  \label{eq:mcmcu} 
\end{eqnarray}
We note that this approach allows us to compute population parameters for the entire vector $\bm{u}$, which includes the vector $\bm{y}$, which is not observed, as well as the vector $\bm{x}$, for which a noisy measurement, $\hat{\bm{x}}$, is available. While it is clear that parameters related to $\bm{y}$ are useful, in some instances it is also useful to estimate population parameters related to $\bm{x}$. A case in point is the image classification problem considered in the following section. In this problem $\hat{\bm{x}}$ represents the input image and $\bm{y}$ represents its label. Here computing parameters associated with $\bm{y}$ provide information about the label for an image. In addition, computing $\bm{x}^{\rm map}$ is useful since large values of the quantity $\|\bm{x}^{\rm map} - \hat{\bm{x}} \|$, which measures the distance between the mode of the posterior distribution and the input image, are strongly correlated with input images that lie outside of the range of the prior, thus enabling the detection of out of distribution (OOD) samples. 

\paragraph{Summary} We have described a method for probing the posterior distribution in two broad classes of problems when the prior is defined by a GAN. The steps of our algorithm are: (1)  Train a GAN using the sample set $\mathcal{S}$ to learn the prior distribution. (2) Reformulate the posterior distribution in the latent space of the GAN.(3) Run a Markov chain Monte Carlo algorithm to generate samples from this low-dimensional posterior distribution. (4) Use MCMC-generated samples to compute population parameters that quantify the uncertainty in the inference.

In the following section we apply  the above algorithm to a broad class of problems where we draw inferences from noisy measurements and quantify uncertainty in these inferences. Wherever possible, we compare our predictions with related methods and/or benchmark solutions and also highlight the role of uncertainty quantification in downstream tasks. In Appendix \ref{sec:appmap}, we also derive  a computationally efficient approach for estimating the MAP for the posterior density of the latent vector under the assumptions of Gaussian noise and prior.

\section{EXPERIMENTS}
In this section we apply our method to the two broad classes considered earlier. One where the forward map is known and another where it is inferred from data. In each case we apply our method to determine important population parameters that include $\bm{y}^{\rm mean}$, ${\rm var} (\bm{y})$ and $ \| \hat{\bm{x}} - \bm{x}^{\rm map}\|$. Thereafter, we use these to answer important questions like: Is the input to the inference problem consistent with the prior data it was trained on? Do we have confidence in the inference? How do we utilize this knowledge in order to design the next measurement.


In all cases we use a Wasserstein GAN-GP (\cite{gulrajani2017improved}) to learn the prior density (architecture described in Appendix \ref{sec:arch}). We also ensure that the target images are not chosen from the set used to train the GAN. We sample from the posterior using Hamiltonian Monte Carlo (\cite{Brooks2012}) and implement it using Tensorflow-probability library. We use initial step size of 1.0 for HMC and adapt it following (\cite{Andrieu2008}) based on the target acceptance probability. We use 64k samples with burn-in period of 0.5. We select these parameters to ensure convergence of chains. Using the HMC sampler we compute the population parameters of interest.




\begin{table}
\renewcommand{\arraystretch}{1.5}
\caption{Comparison of different hybrid models. Arrows indicate which direction is better.}
\label{tab:hybrid_compare}
\centering
\begin{adjustbox}{width={\linewidth}}
\begin{tabular}{c l c c c}
\toprule
&  Configuration /  & \multicolumn{2}{c}{\bfseries MNIST} & \bfseries NotMNIST\\
\cline{3-5}
& Rejection rule  & Acc $\uparrow$ & FPR $\downarrow$ & Entropy $\uparrow$ \\
\midrule
\cite{Nalisnick2019} & \multirow{2}{*}{log $p(x)$}  &  \multirow{2}{*}{95.99 \%} & \multirow{2}{*}{-} & \multirow{2}{*}{\textbf{2.300}}  \\
($\lambda$ = 10.0/D) & & & & \\
\hline
\multirow{2}{*}{\cite{Chen2019}} 
& Coupling & 95.42\% & - & - \\
\multirow{2}{*}{($\lambda = 1$)}
& + 1 $\times$ 1 Conv & 94.22\%& - & - \\
& Residual & 98.69\%& - & - \\
\hline
\multirow{2}{*}{Ours} &
$\|\hat{\bm{x}} - \bm{x}^{\rm map}\|$ & 96.81\% & 0 & $\textbf{2.300}$ \\
&  +  $\|$ var($\bm{y}$) $\|$  & $\textbf{99.57\%}$ & 0.064 & $\textbf{2.300}$ \\
\bottomrule
\end{tabular}
\end{adjustbox}
\end{table}

\subsection{IMAGE CLASSIFICATION}
This problem belongs to the second class, where the forward map is not known and must be learnt from data. The objective of this task is to infer the label $\bm{y}$ along with its uncertainty for a given input image $\hat{\bm{x}}$. This predictive uncertainty estimation is crucial in deep learning applications where high-stakes decisions are made based on the output of a model (\cite{Kahn2017}). 
It has been shown that in real-world scenarios, where a model might encounter inputs that are anomalous to its training data distribution, many models produce overconfident predictions (\cite{Lakshminarayanan2016}) raising serious concerns about AI safety (\cite{Amodei2016}). In this situation, it is desirable that such out-of-distribution (OOD) data points are detected upfront before making any prediction. A useful probabilistic predictive model should therefore flag all OOD data points, maintain high levels of accuracy on in-distribution data points, and provide a measure of confidence in its predictions.
In order to achieve this goal, we compute three different quantities: $ \| \hat{\bm{x}} - \bm{x}^{\rm map}\|$ for OOD detection, $\bm{y}^{\rm mean}$ for prediction, and ${\rm var} (\bm{y})$ as a measure of confidence in the prediction. 

We consider the MNIST database of hand-written digits and use 55k images and the corresponding labels to train a  WGAN-GP. Thereafter, we use the MNIST test set to test the performance of our algorithm for in-distribution data, and NotMNIST test set for OOD data. Our approach of learning and inferring the joint distribution is closely related to hybrid models and hence we compare our performance against the most recent hybrid models in Table \ref{tab:hybrid_compare}. 

We determine whether a given test image is OOD based on a rejection rule. If this condition is satisfied then following \cite{Nalisnick2019} we set the probability of each label to be equal. We then quantify the performance of the rejection rule by reporting the average entropy of the labels for all test samples from the OOD set and the false positive rate (FPR = \# of in-distribution samples rejected as OOD/ \# of in-distribution samples). Thereafter, for all in-distribution samples that are correctly identified, we report the accuracy of predicting the label, which is determined from $\bm{y}^{\rm mean}$. We consider two rejection rules: $\|\hat{\bm{x}} - \bm{x}^{\rm map}\| > c_1$, and $\|\hat{\bm{x}} - \bm{x}^{\rm map}\| + \|$var($\bm{y}$)$\| > c_2$. 

\begin{figure}[h]
\begin{center}
\includegraphics[width=0.9\linewidth]{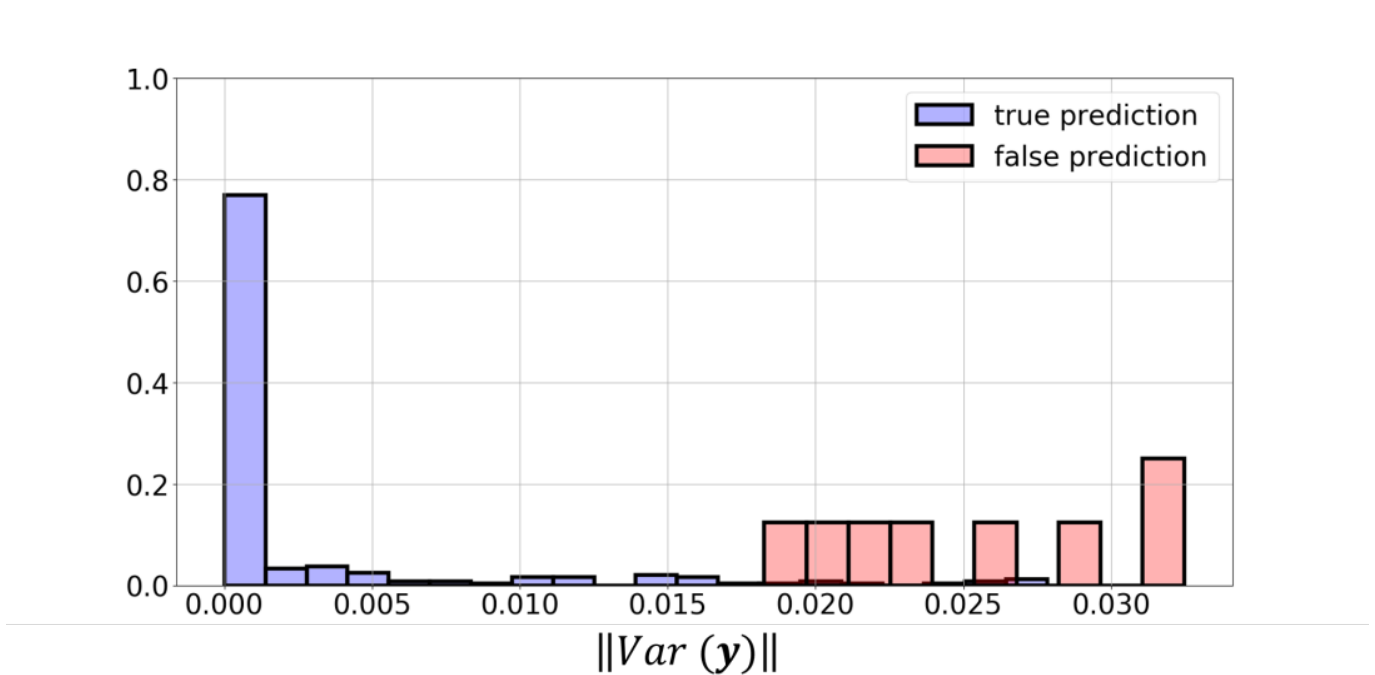} 
\end{center}
\caption{\label{fig:varY_hist} Histogram of $\|$var($\bm{y}$)$\|$ for MNIST dataset.} 
\end{figure}

The performance of $\|\hat{\bm{x}} - \bm{x}^{\rm map}\| > c_1$  rejection rule can be discerned in Figure \ref{fig:intro}, where we observe that it perfectly segregates the in-distribution and OOD samples. This is also apparent in Table \ref{tab:hybrid_compare}, where it yields zero FPR and maximum entropy. Its accuracy for the in-distribution samples is also quite high. This accuracy can be further improved by using the combined rejection rule $\|\hat{\bm{x}} - \bm{x}^{\rm map}\| + \|$var($\bm{y}$)$\| > c_2$, since it rejects some incorrectly labeled in-distribution samples with high variance as OOD. However, this comes at the cost of a slightly higher FPR. The usefulness of $\|$var($\bm{y}$)$\|$ as a predictor of accuracy is evident in Figure \ref{fig:varY_hist}, where we observe that most correctly labeled samples have low variance (avg. value = 0.0026 $\pm$ 6e-3) when compared with their incorrectly labeled counterparts (avg. value = 0.025 $\pm$ 5e-3). 


In Table \ref{tab:hybrid_compare} we compare the performance of our GAN-based approach with two hybrid invertible flow-based models \cite{Nalisnick2019,Chen2019}. The explicit nature of these models allows the joint density to be decomposed into generative and discriminative components, enabling a way to explore the generative-discriminative trade-off by introducing a weighted likelihood objective with a scaling parameter $\lambda$. Values of  $\lambda<1$ favor discriminative performance, while $\lambda>1$ favors generative performance. In this context our approach may be regarded as one  where the generative and discriminative components are equally weighted, and is therefore close to the choice $\lambda =1$. Given this, in Table \ref{tab:hybrid_compare}, we have compared our approach with hybrid models where $\lambda \approx 1$. We note that with the $\|\hat{\bm{x}} - \bm{x}^{\rm map}\| \le c_1$ rejection rule our model performs competitively with both the scaled (\cite{Nalisnick2019}) and the un-scaled hybrid models (\cite{Chen2019}) for both in-distribution and OOD datasets. With the $\|\hat{\bm{x}} - \bm{x}^{\rm map}\| + \|$var($\bm{y}$)$\| > c_2$ rejection rule it outperforms both hybrid models giving maximum accuracy and entropy but with non-zero FPR.

\subsection{IMAGE INPAINTING} \label{sec:img_recovery_mnist}
Image inpainting also belongs to second class of problems, where the forward map is unknown and has to be learnt from data. In this case the quantity to be inferred ($\bm{y}$) is the occluded region of an image, and the measurement ($\hat{\bm{x}}$) is the noisy version of its visible portion. The goal is to recover the entire image ($\bm{u} = [\bm{x}, \bm{y}]$).  
While there has been great interest in recent years in developing efficient deep learning-based image inpainting algorithms (\cite{Yu_2018_CVPR}), most of it has focused on deterministic algorithms that lack the ability to quantify uncertainty in a prediction. In contrast, we use the algorithm in Section \ref{S:2} to perform probabilistic image inpainting.

\begin{figure}[h]
\begin{center}
\includegraphics[width=0.7\linewidth]{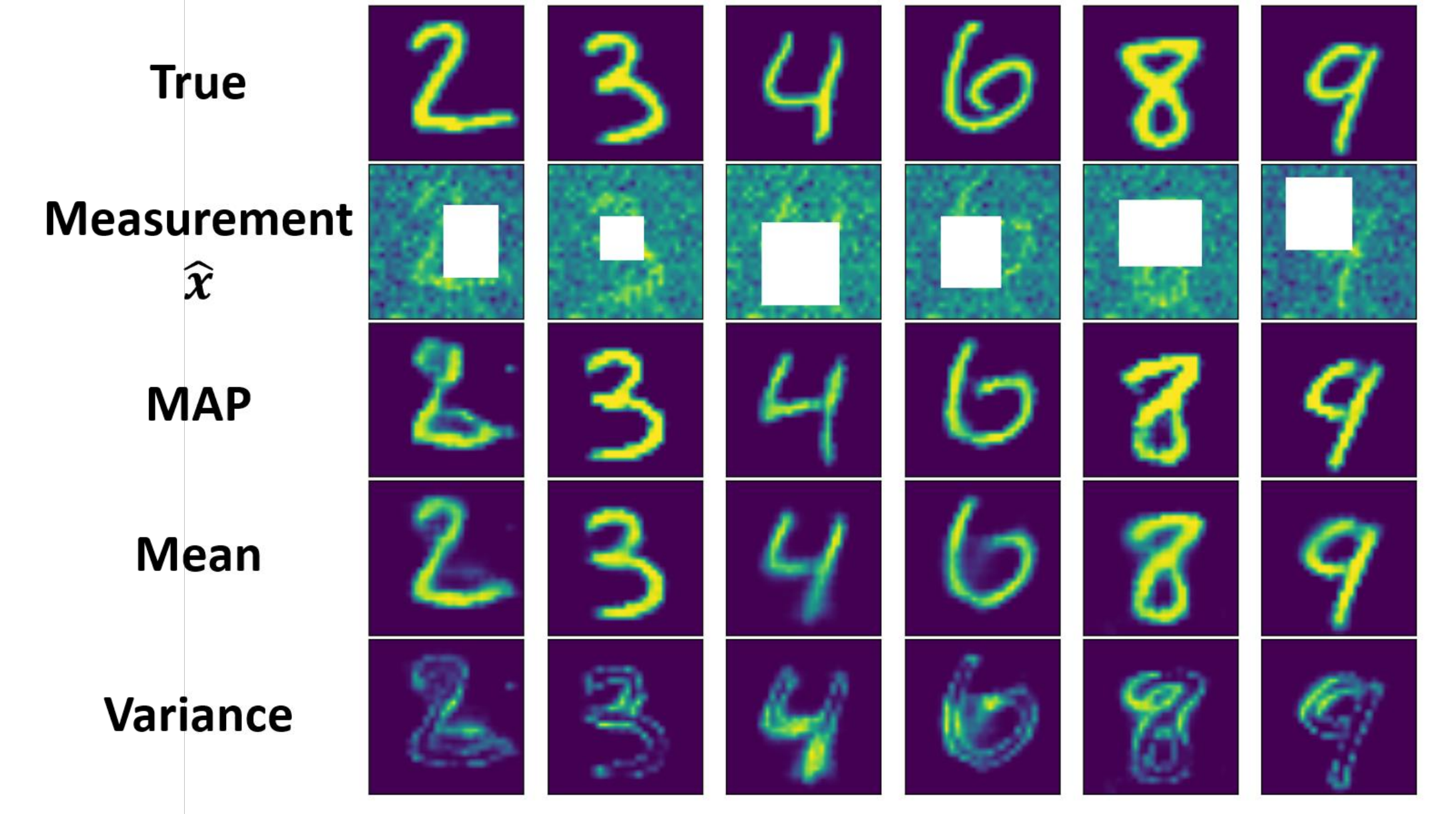} 
\end{center}
\caption{\label{fig:mnist1} Estimate of the MAP, mean and pixel-wise variance from noisy occluded images using the proposed method. The variance is peaked at the occluded region. }  
\end{figure}

We consider MNIST dataset and use 55k images to train a WGAN-GP. We generate measurements by selecting an image from the test set, occluding a significant region and then adding Gaussian noise. With this as input we use our algorithm to generate samples from the posterior distribution of the entire image (both occluded and retained regions). From these samples we evaluate the relevant population parameters, $\bm{u}^{\rm map}$, $\bm{u}^{\rm mean}$, and var($\bm{u}$). These results are shown in Figure \ref{fig:mnist1} and indicate that the map and mean images are close to the true image, even in the presence of significant occlusion and noise. The image of pixel-wise variance reveals that we are most uncertain along the boundaries of the digits and around the occlusion window. 

\begin{figure}[h]
\begin{center}
\includegraphics[width=0.7\linewidth]{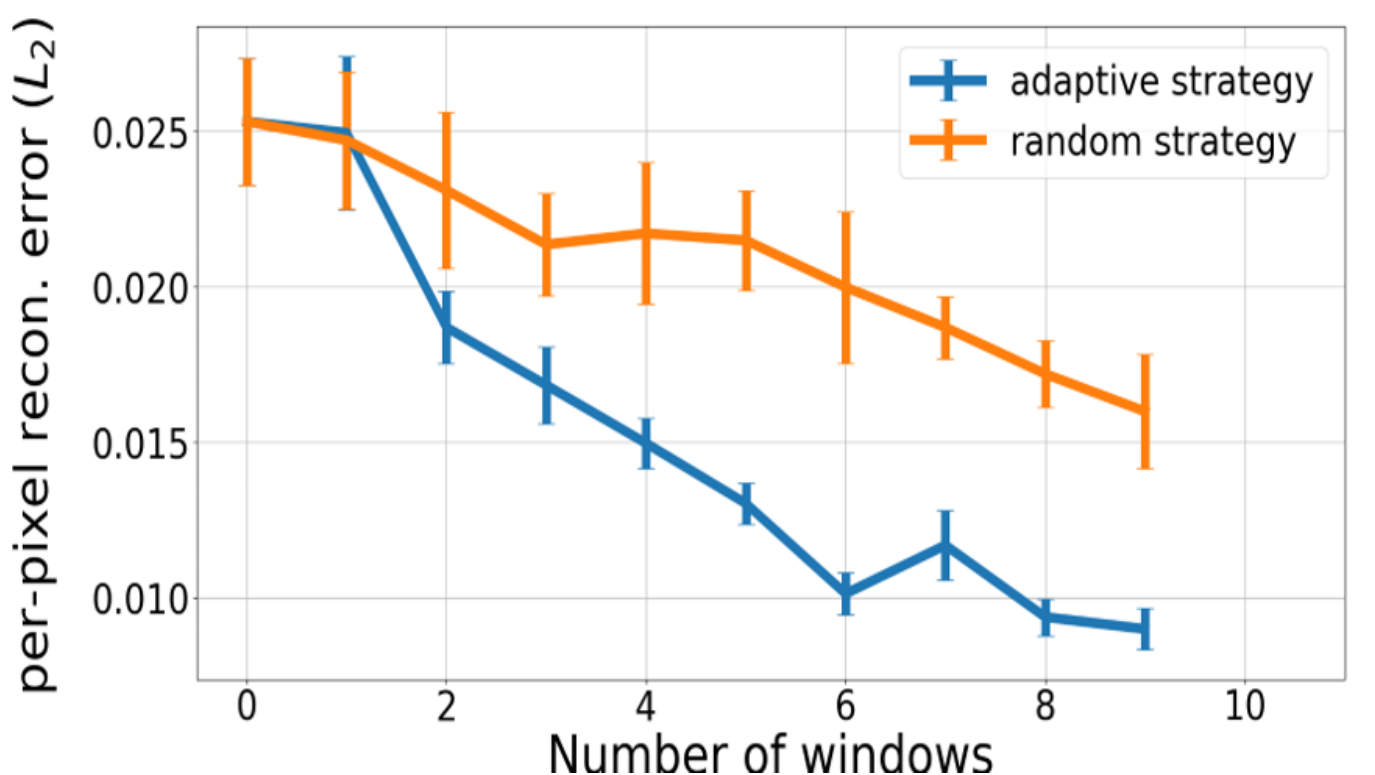} 
\end{center}
\caption{\label{fig:oed} Average reconstruction error (with 95\% confidence interval) as a function of number of windows for variance-driven (adaptive) and random sampling strategies.} 
\end{figure}

In Figure \ref{fig:intro}, we demonstrate how uncertainty  may be used in active learning/design of experiment, \dpadd{where the goal is to determine the optimal location for a measurement}. We begin with an input where the entire image is occluded and in every subsequent step, \dpadd{we} allow for a small 7$\times$7 pixel window to be revealed. We select the window with the largest average pixel-wise variance. As the iterations progress, the MAP estimate converges to the true digit, and the variance decreases. In about 4 iterations we arrive at a very good guess for the digit. The performance of this approach is quantified in Figure \ref{fig:oed}, where we have plotted reconstruction error versus the number windows for this strategy, and a strategy where the subsequent window is selected randomly. The variance-driven strategy consistently performs better. 
\dpadd{We are not aware of any other methods for computing uncertainty in recovered images that have been applied to drive an active learning task in image inpainting. While methods based on dropout (\cite{Kendall2019}) or variational inference (\cite{Kohl2018a}) could be extended to accomplish this, this has not been done thus far.}

\begin{figure}[h]
\begin{center}
\includegraphics[width=\linewidth]{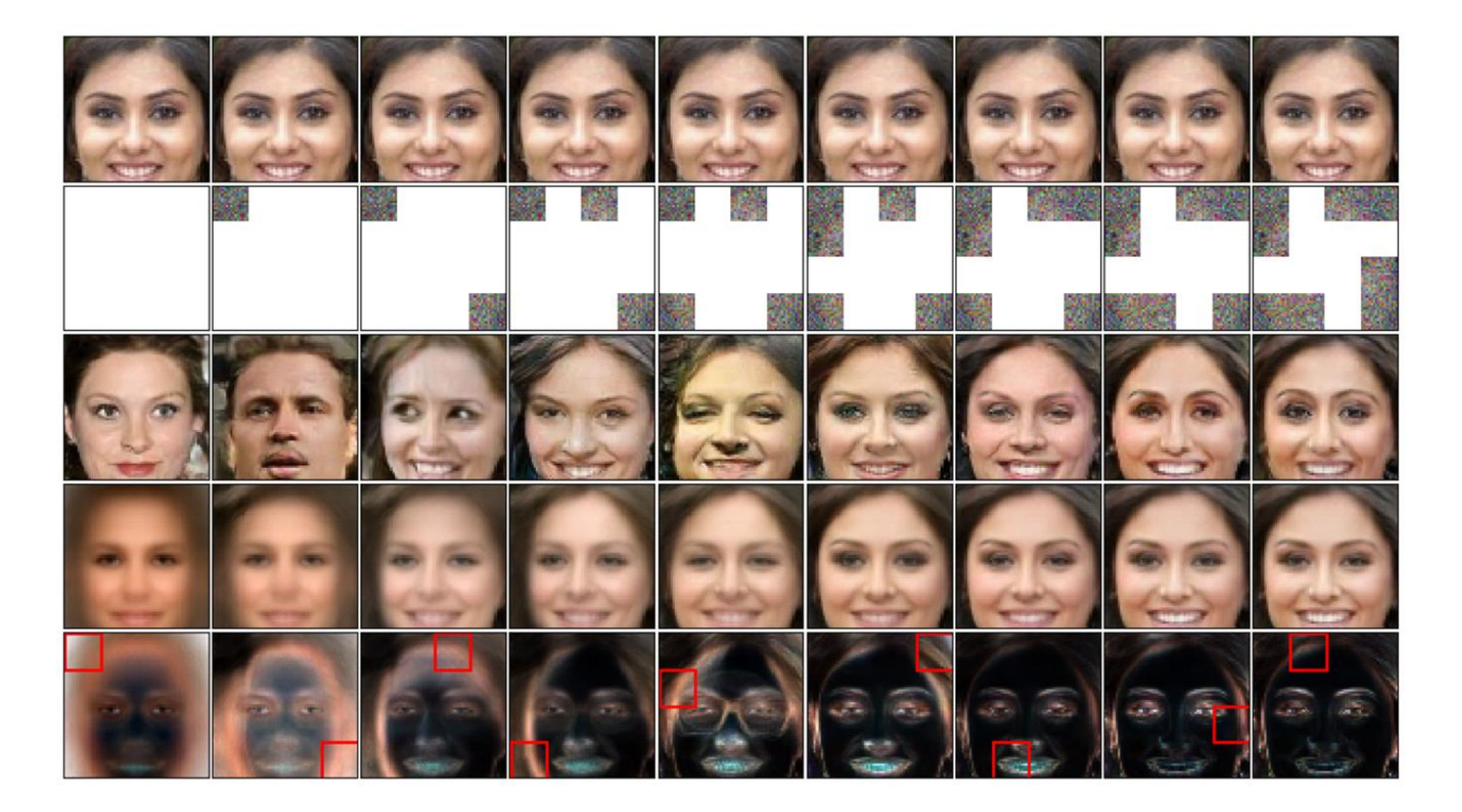} 
\end{center}
\caption{\label{fig:celeb1} CelebA dataset: Estimate of the MAP, mean and variance from the limited view of a noisy image (2nd row) of a true image (1st row) using the variance-driven adaptive learning strategy.} 
\end{figure}

Results for the variance-based window selection strategy applied to the CelebA dataset are shown in Figure \ref{fig:celeb1}. We observe that the algorithm produces realistic images at each iteration; however, the initial variance is large \aoadd{indicating large uncertainty}. As more windows are sampled using the active learning  strategy, the variance reduces and by the 7th iteration a good approximation of the true image is obtained, even though only a small, noisy portion is revealed. Additional results for this dataset are discussed in Appendix \ref{ref:append:celeba}.

\subsection{A PHYSICS-DRIVEN INFERENCE PROBLEM}
We now consider a problem from Class 1, where the forward map is known. In particular,  we consider the problem of inverse heat conduction, where the goal is to infer the initial temperature distribution (at $t=0$) in a domain given a noisy measurement of temperature  at later time ($t=1$) and the thermal conductivity of the material. The forward map is the solution of the time-dependent heat conduction problem with uniform conductivity, $\kappa = 0.64$, in a square domain of length $L = 2\pi$ with Dirichlet boundary conditions. This operator maps the initial temperature field ($\bm{y}$) to the temperature field at later time ($\bm{x}$). Its  discrete version is obtained by discretizing the time-dependent linear heat conduction equation using the central difference scheme in space and backward Euler scheme in time. Much like a blurring kernel, the forward operator smooths the initial temperature distribution, and the extent of smoothing increases with $\kappa \times t$. 

We consider a family of initial temperatures where the background is zero, and the temperature on a rectangular sub-domain varies linearly from  2 units on the left edge to 4 units on the right edge. This distribution is parameterized  by four parameters, $\{\xi_i\}_{i=1}^4$, which are the coordinates of the lower left and upper right corners of the rectangular region. The sample set $\mathcal{S}$ is created by sampling each parameter from a uniform distribution  and is used to train the GAN  prior. The posterior distribution is sampled using the HMC sampler. 

\begin{figure}[h]
\begin{center}
\includegraphics[width= 0.95\linewidth]{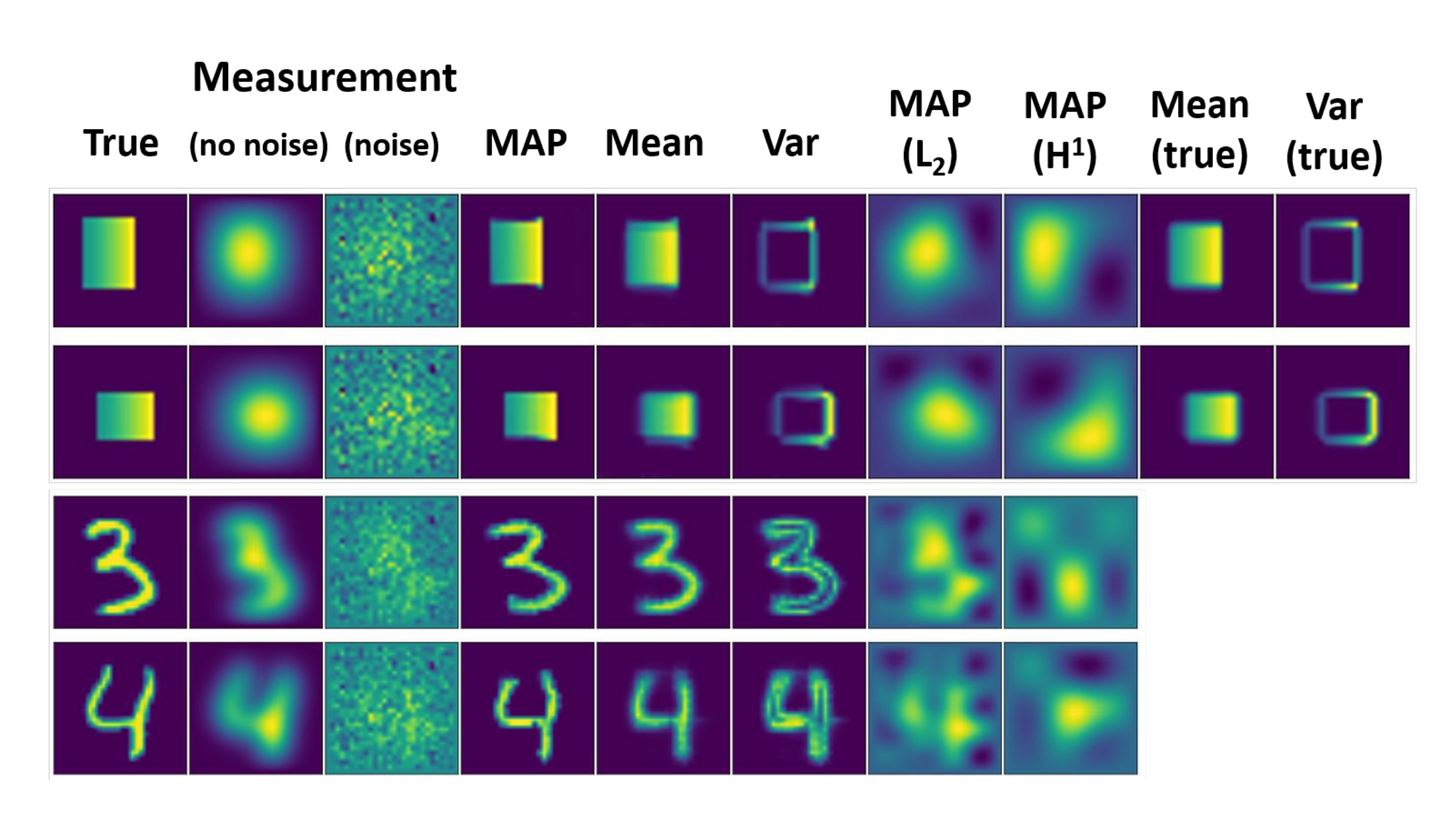} 
\end{center}
\caption{\label{fig:ic1} From left to right: (1) true initial temperature, (2) temperature at $t=1$, (3) noisy version temperature used as measurement, (4), (5), (6) MAP, mean and pixel-wise variance estimates using GAN priors, (7) and (8) MAP estimates using $L_2$ and $H^1$ Gaussian priors, (9) \& (10) true MAP and variance.} 
\end{figure}

In the top two rows of Figure \ref{fig:ic1}, we have plotted the true initial condition, the noise-free temperature at $t=1$, and the noisy temperature measurement ($\sigma_x = 1$) used as input in the GAN-based prior approach. The corresponding MAP, mean and pixel-wise variance estimated by the MCMC approximation are shown next. We observe that the MAP is very close to the true initial temperature distribution and the variance is concentrated along the edges of the rectangle where the uncertainty is the largest. In the following columns we have plotted the MAP estimate obtained assuming $L_2$ and $H^1$ Gaussian priors, which are often used to solve these types of problems, and are clearly much less accurate. 

For this problem the ``true'' posterior  can be reduced to the 4-dimensional space of parameters, and sampled by generating initial conditions corresponding to the values of these parameters. A simple Monte-Carlo approximation can be performed to compute the mean and the pixel-wise variance for the true posterior (last two columns of Figure \ref{fig:ic1}). By comparing these with the mean and the pixel-wise variance (columns 5 \& 6) estimated by the GAN-based prior, we conclude that GAN-based posterior has converged to the true posterior.  

In the bottom rows of Figure \ref{fig:ic1}, we plot similar results for initial conditions generated from the MNIST database when the GAN prior was also trained on the MNIST database. The measurement is made at $t=0.2$. Once again we observe that mean and the map estimated by our approach is very close to the true initial condition, while the MAP solutions obtained from the $L_2$ and $H^1$ priors are inaccurate. The pixel-wise variance illustrates the uncertainty in determining the boundary of the digits.



\subsection{IMAGE DENOISING}
\begin{figure}[h]
\begin{center}
\includegraphics[width=0.85\linewidth]{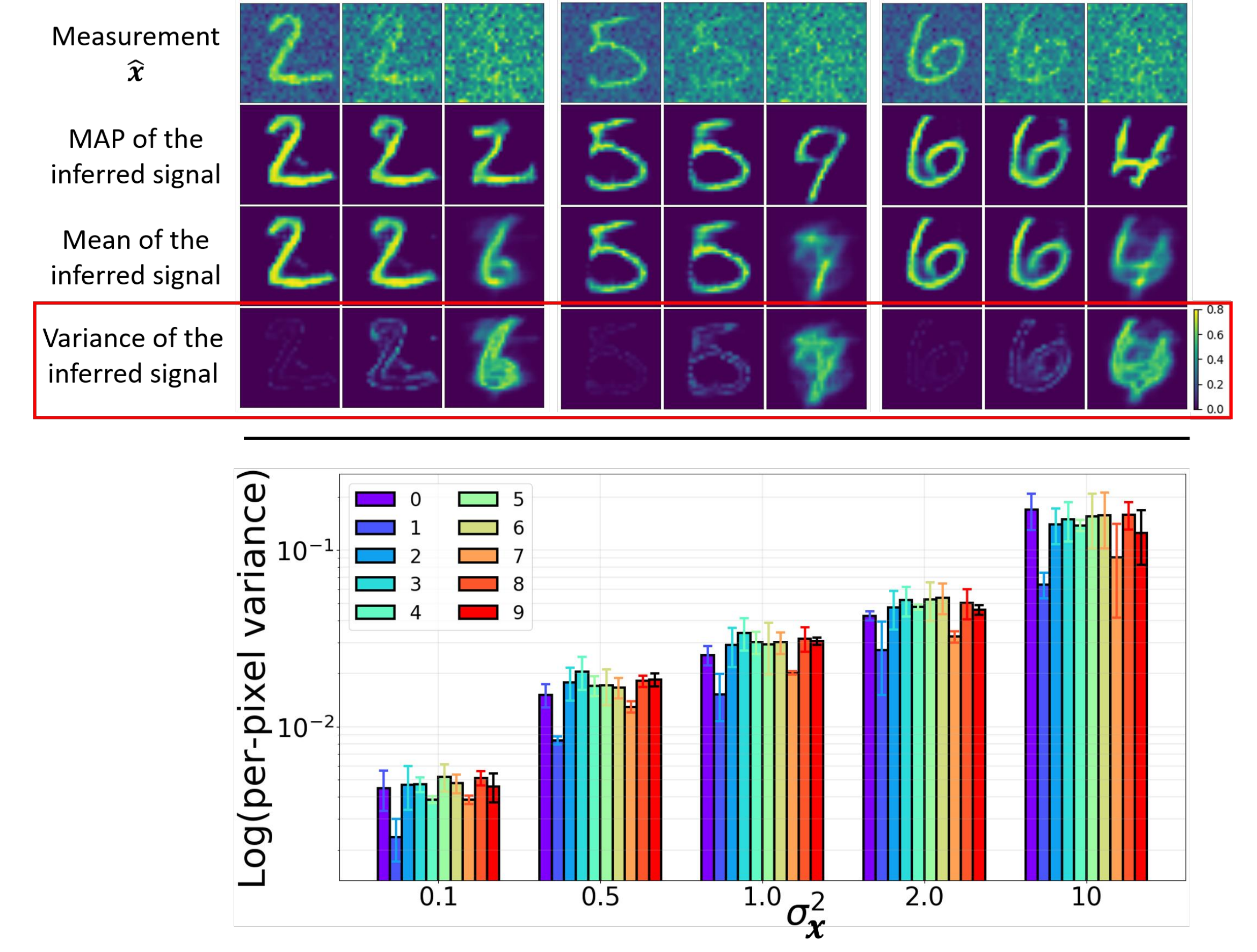} 
\end{center}
\caption{\label{fig:denoise} Top panel: Estimate of the MAP, mean and pixel-wise variance from a noisy image using the proposed method. In the first three panels $\bm{\sigma_x} = 0.1,1, \& 10$, when moving from left to right. Bottom panel: Average variance per pixel ($\overline{var(\bm{y})}$) in a reconstructed image as a function of variance of noise for 10 MNIST digits (along with 95\% confidence interval).} 
\end{figure}

As another example of a problem where the forward map is known (Class 1 problem) we consider image denoising. Here the forward map is the identity, the measured data is the noisy image and the inferred field is its de-noised version. We consider the MNIST dataset and use 55k images to train the GAN. 
We add Gaussian noise with zero mean and specified variance ($\sigma_x$) \dpadd{to the test images} and use these as  \dpadd{measurements} to recover the distribution of likely images using the MCMC approach. In Figure \ref{fig:denoise}, we have plotted the noisy input image, the MAP estimate, and the pixel-wise mean and variance. For low and medium noise levels ($\sigma_x = 0.1, 1$), we are able to recover the original image with good accuracy, the pixel-wise variance is small overall, and is largest around the boundary of the recovered digit. For the highest noise level ($\sigma_x = 10$), \dpadd{however,} the image recovered by the MAP is incorrect in 2/3 cases, and would be misleading if viewed by itself. However, when viewed in conjunction with the estimated variance, which is large, it is clear that the confidence in the \dpadd{inference} is small \dpadd{and the inferred image ought not be trusted for downstream tasks}. The dependence of the average per-pixel variance in the recovered image on the variance of noise in the measured image is shown in Figure \ref{fig:denoise}, and it increases with noise. Additional results for CelebA are discussed in Appendix C.2

\vspace{-0.2 cm}
\section{CONCLUSIONS}

The ability to quantify the uncertainty in an inference problem is useful in developing confidence in that inference, identifying measurements that are outliers, and in designing strategies to improve the confidence. 
In this paper we have described how this may be accomplished when solving a Bayesian inference problem by using GANs as priors. Since GANs can learn complex distributions of a wide variety of fields from their samples, this approach can be applied to a  range of problems in computer vision and physics-driven inference. This includes those where the operator that maps the inferred field to measurement is known (so-called inverse problems) and those where this map is not known and must be inferred from data. It derives its efficiency by mapping the posterior distribution to the latent space, whose dimension is often much smaller than that of the inferred field. We have presented applications of this approach to image classification, image inpainting, image denoising and physics-driven inverse problems. 



\bibliographystyle{plainnat} 
{\bibliography{bib_uai2020}}

\clearpage

\appendix
\section{Weak convergence of the prior density \label{sec:appweak}}

Let the generator of the Wasserstein GAN be given by $\bm{g}(\bm{z};\bm{\theta})$, where $\bm{z} \in \mathbb{R}^M$ is the latent vector, and $\bm{\theta} \in \mathbb{R}^{N_\theta}$ is the vector of weights. The vector $\bm{z}$ is selected from the distribution $p_Z(\bm{z})$. Note that $\bm{g}: \mathbb{R}^M \to \mathbb{R}^N$. 

Let the discriminator of the GAN be given by $d(\bm{y};\bm{\phi})$, where $\bm{y} \in \mathbb{R}^N$, and $\bm{\phi} \in \mathbb{R}^{N_\phi}$ be the vector of weights. Note that $d: \mathbb{R}^N \to \mathbb{R}(0,1)$. 

Assume that the GAN is trained using a set of samples of $\bm{y}$, drawn from $p_Y(\bm{y})$. 

Then under the following assumptions:
\begin{enumerate}
    \item The stationarity conditions for the adversarial loss function are satisfied.
    \item In the limit of infinite capacity ($N_\phi \to \infty $), the set of basis functions obtained by taking the derivative of the discriminator with respect to its weights forms a complete basis in $L^\infty(\Omega_y)$
\end{enumerate} 
For a sufficiently smooth $m(\bm{y})$, we prove that \begin{eqnarray}
 |\expec_{\bm{y} \sim p_Y} [m(\bm{y})]| =  | \expec_{\bm{z} \sim p_Z} [ m (\bm{g}(\bm{z}))]|.
\end{eqnarray}
That is we establish the weak convergence of the density obtained by using a GAN as a prior to the true prior density.

%
%
\paragraph{Proof.} For the loss function, consider 
\begin{eqnarray}
\label{eq:defl} L (\bm{\theta},\bm{\phi}) &=& \expec_{\bm{y} \sim p_Y } [ \rho( 1- d(\bm{y};\bm{\phi}))]  \nonumber \\
& & +  \expec_{\bm{z} \sim p_Z} [\rho(d(\bm{g}(\bm{z}; \bm{\theta}); \bm{\phi}))]. 
\end{eqnarray}
Here $\rho$ is a monotone real-valued function which defines the GAN family being analyzed. For example, for the Wasserstein GAN, $\rho(\xi) = \xi$.

The optimal values of the weights are given by 
\begin{eqnarray}
\bm{\theta}^{*}, \bm{\phi}^* =  \amax_{\bm{\theta}} (\amin_{\bm{\phi}} (L (\bm{\theta},\bm{\phi}))).
\end{eqnarray}

The necessary conditions for these optimal values are 
\begin{eqnarray}
\label{eq:dphi1} \frac{\partial L (\bm{\theta}^*,\bm{\phi}^*) }{\partial \bm{\phi}} &=& \bm{0} \\
\label{eq:dtheta1} \frac{\partial L (\bm{\theta}^*,\bm{\phi}^*) }{\partial \bm{\theta}} &=& \bm{0}.
\end{eqnarray}

Using the definition of the loss function (\ref{eq:defl}) in (\ref{eq:dphi1}), we have 
\begin{eqnarray}
\label{eq:dphi2} \expec_{\bm{x} \sim p_Y} [ \rho'( 1- d(\bm{y};\bm{\phi}^*)) \frac{\partial d}{\partial \bm{\phi}} (\bm{y};\bm{\phi}^*) ] = \qquad \qquad \nonumber \\
\qquad \expec_{\bm{z} \sim p_Z} [\rho'(d(\bm{g}(\bm{z}; \bm{\theta}^*); \bm{\phi}^*)) \frac{\partial d} {\partial \bm{\phi}} (\bm{g}(\bm{z}; \bm{\theta}^*);\bm{\phi}^*)]. 
\end{eqnarray}

Similarly, using (\ref{eq:defl}) in (\ref{eq:dtheta1}), we have 
\begin{eqnarray}
\label{eq:dtheta2}  \expec_{\bm{z} \sim p_Z} [\rho'(d(\bm{g}(\bm{z}; \bm{\theta}^*); \bm{\phi}^*)) \frac{\partial d} {\partial \bm{x}} (\bm{g}(\bm{z}; \bm{\theta}^*);\bm{\phi}^*) \cdot \nonumber \\
\frac{\partial \bm{g}}{\partial \bm{\theta}} (\bm{z}; \bm{\theta}^*) ] = \bm{0}.
\end{eqnarray}

For the Wasserstein GAN, $\rho(\xi) = \xi$ and $\rho'(\xi) = 1$. As a result (\ref{eq:dphi2}) reduces to 
\begin{eqnarray}
\label{eq:dphi3} \expec_{\bm{y} \sim p_Y} [ \frac{\partial d}{\partial \bm{\phi}} (\bm{y};\bm{\phi}^*) ] &=& \expec_{\bm{z} \sim p_Z} [ \frac{\partial d} {\partial \bm{\phi}} (\bm{g}(\bm{z}; \bm{\theta}^*);\bm{\phi}^*)],
\end{eqnarray}
and (\ref{eq:dtheta2}) reduces to,
\begin{eqnarray}
\label{eq:dtheta3}  \expec_{\bm{z} \sim p_Z} [ \frac{\partial d} {\partial \bm{y}} (\bm{g}(\bm{z}; \bm{\theta}^*);\bm{\phi}^*) \cdot \frac{\partial \bm{g}}{\partial \bm{\theta}} (\bm{z}; \bm{\theta}^*) ] &=& \bm{0}.
\end{eqnarray}

Let $w_a(\bm{y}) \equiv \frac{\partial d}{\partial \phi_a} (\bm{y};\bm{\phi}^*)$, then for $, a = 1, \cdots, N_\phi$.  (\ref{eq:dphi3}) implies 
\begin{eqnarray}
\label{eq:dphi4} \expec_{\bm{y} \sim p_Y} [ w_a (\bm{y}) ] &=& \expec_{\bm{z} \sim p_Z} [ w_a (\bm{g}(\bm{z}; \bm{\theta}^*))]. 
\end{eqnarray}
As $N_\phi \to \infty$, this equation implies that 
the push forward of the measure in the latent space under the function $\bm{g}(\bm{z})$ weakly converges to the measure associated with distribution of $\bm{y}$. We note with increasing number of weights in the discriminator, the relation above is required to hold for an increasing number of test functions, $w_a$. In addition, we have implicitly assumed that the generator is rich enough, that is it has enough weights/layers, such that this relation can actually be satisfied. To make this clear consider the extreme case of a generator with a single weight; in this case there is no way that (\ref{eq:dphi4}) will be satisfied for a large number $N_\phi$. Thus in order for this relation to hold for a large $N_\phi$, we must also provide the generator with a large $N_\theta$. 

Let $\mathcal{V} \equiv {\rm span} \{w_a(\bm{y}), a = 1, \cdots, N_\phi\}$. Then from (\ref{eq:dphi4}) for any $v \in \mathcal{V}$, we have 
\begin{eqnarray}
\label{eq:dphi5} \expec_{\bm{y} \sim p_Y} [v (y) ] &=& \expec_{\bm{z} \sim p_Z} [ v (\bm{g}(\bm{z}))].
\end{eqnarray}
In the equation above, and hereafter, we have suppressed the arguments $\bm{\theta}^*$ and $\bm{\phi}^*$ for ease of notation, implicitly assuming that the relations hold at the optimal values of weights. 
Now consider a sufficiently smooth function $m(\bm{y})$ which defines the point estimate we wish to compute, and let $\bar{m}(\bm{y})$ be its $L^\infty$ projection on to $\mathcal{V}$. That is,
\begin{eqnarray}
\bar{m}(\bm{y}) = \amin_{v \in \mathcal{V}} \| m(\bm{y}) - v(\bm{y}) \|_{L^{\infty}(\Omega_y)}.
\end{eqnarray}
and let $\epsilon = \| m(\bm{y}) - \bar{m}(\bm{y}) \|_{L^{\infty}(\Omega_y)}$. 
Given the assumption that the functions $w_a$ form a complete basis in $L^{\infty}(\Omega_y)$, we note that as $N_\phi \to \infty$, $\epsilon \to 0$.

Now consider the difference, 
\begin{eqnarray}
& & |\expec_{\bm{y} \sim p_Y} [m(\bm{y})] - \expec_{\bm{z} \sim p_Z} [ m (\bm{g}(\bm{z}))]| \nonumber \\
& & \le 
|\expec_{\bm{y} \sim p_Y} [m(\bm{y})] - \expec_{\bm{y} \sim p_Y} [\bar{m}(\bm{y})] \nonumber \\
& & \qquad + \expec_{\bm{z} \sim p_Z} [ \bar{m} (\bm{g}(\bm{z}))]- \expec_{\bm{z} \sim p_Z} [ m (\bm{g}(\bm{z}))]| \nonumber \\
& & \le |\expec_{\bm{y} \sim p_Y} [m(\bm{y})- \bar{m}(\bm{y})]| \nonumber \\
& & \qquad + | \expec_{\bm{z} \sim p_Z} [ m (\bm{g}(\bm{z}))- \bar{m} (\bm{g}(\bm{z}))]| \nonumber \\ 
& &\le \expec_{\bm{y} \sim p_Y} [\epsilon] + | \expec_{\bm{z} \sim p_Z} [\epsilon] \nonumber \\
& & = 2 \epsilon.
\end{eqnarray}
In the equation above, the first inequality is obtained by recognizing that $\bar{m}(\bm{y}) \in \mathcal{V}$ and using (\ref{eq:dphi5}), the second inequality is a consequence of the triangle inequality and the third is due to the definition of $\epsilon$. Now in the limit $N_\phi \to \infty$, $\mathcal{V}$ tends to a complete basis, therefore $\epsilon \to 0$ and we have the desired result.

\section{Expression for the maximum a-posteriori estimate \label{sec:appmap}}


The techniques described in Section 2.1 focus on sampling from the posterior distribution and computing approximations to population parameters. These techniques can be applied in conjunction with any distribution used to model noise and the latent space vector; that is, any choice of $p_\eta$ (likelihood) and $p_Z$ (prior). In this section we consider the special case when Gaussian models are used for noise and the latent vector. In this case, we can derive a simple optimization algorithm to determine the maximum a-posteriori estimate (MAP) for $p^{\rm post}_Z(\bm{z}|\bm{x})$. This point is denoted by $\bm{z}^{\rm map}$ in the latent vector space and represents the most likely value of the latent vector in the posterior distribution. It is likely that the operation of the generator on $\bm{z}^{\rm map}$, that is $\bm{g}(\bm{z}^{\rm map})$, will yield a value that is close to $\bm{y}^{\rm map}$, and may be considered as a likely solution to the inference problem. 

We consider the case when the components of the latent vector are iid with a normal distribution with zero mean and unit variance. This is often the case in many typical applications of GANs. Further, we assume that the components of noise vector are defined by a normal distribution with zero mean and a covariance matrix $\bm{\Sigma}$. Using these assumptions in (\ref{eq:pzposty}), we have 
\begin{eqnarray}
\resizebox{0.9\linewidth}{!}{%
$p^{\rm post}_Z(\bm{z}|\bm{x}) \propto \exp{\Big( -\frac{1}{2}\overbrace{\big( | \bm{\Sigma}^{-1/2} (\hat{\bm{x}} - \bm{f}(\bm{g}(\bm{z})))|^2  + |\bm{z}|^2 \big)}^{\equiv r(\bm{z})} \Big)}%
$}.
\label{eq:postZ}
\end{eqnarray}

The MAP estimate for this distribution is obtained by minimizing the negative of the argument of the exponential. That is 
\begin{eqnarray}
\bm{z}^{\rm map} = \amin_{\bm{z}} r(\bm{z}). \label{eq:zmap}
\end{eqnarray}
This minimization problem may be solved using any gradient-based optimization algorithm. The input to this algorithm is the gradient of the functional $r$ with respect to $\bm{z}$, which is given by 
\begin{eqnarray}
\frac{\partial r}{\partial \bm{z}} = \bm{H}^T \bm{\Sigma}^{-1} (\bm{f}(\bm{g}(\bm{z})) - \hat{\bm{x}}) + \bm{z},  \label{eq:gradpi}
\end{eqnarray}
where the matrix $\bm{H}$ is defined as 
\begin{eqnarray}
\bm{H} \equiv \frac{\partial \bm{f}(\bm{g}(\bm{z}))}{\partial \bm{z}} = \frac{\partial \bm{f}}{\partial \bm{y}} \frac{\partial \bm{g}}{\partial \bm{z}}.  
\end{eqnarray}
Here $\frac{\partial \bm{f}}{\partial \bm{y}}$ is the derivative of the forward map $\bm{f}$ with respect to its input $\bm{x}$, and $\frac{\partial \bm{g}}{\partial \bm{z}}$ is the derivative of the generator output with respect to the latent vector. In evaluating the gradient above we need to evaluate the operation of the matrices $\frac{\partial \bm{f}}{\partial \bm{y}}$ and $\frac{\partial \bm{g}}{\partial \bm{z}}$ on a vector, and not the matrices themselves. The operation of $\frac{\partial \bm{g}}{\partial \bm{z}}$ on a vector can be determined using a back-propagation algorithm within the GAN; while the operation of $\frac{\partial \bm{f}}{\partial \bm{y}}$ can be determined by making use of the adjoint of the linearization of the forward operator. 

Once $\bm{z}^{\rm map}$ is determined, one may evaluate $\bm{g}(\bm{z}^{\rm map})$ by using the GAN generator. This represents the value of the field we wish to infer at the most likely value value of latent vector. Note that this is not the same as the MAP estimate of $p^{\rm post}_Y(\bm{y}|\bm{x})$. 

We note that the result derived above applies to Class 1 problems, that is inverse problems. A similar result can also be derived for problems from Class 2, by using (\ref{eq:pzpostu}) as a starting point and repeating the steps outlined above. In this case $\bm{z}^{\rm map}$ is the minimizer of 
'\begin{eqnarray}
r \equiv  | \bm{\Sigma}^{-1/2} (\hat{\bm{x}} - \bm{R}(\bm{g}(\bm{z})))|^2  + |\bm{z}|^2,
\end{eqnarray}
where $\bm{R}$ is the restriction operator. The gradient for this optimization problem is given by 
\begin{eqnarray}
\frac{\partial r}{\partial \bm{z}} = \bm{H}^T \bm{\Sigma}^{-1} (\bm{R}(\bm{g}(\bm{z})) - \hat{\bm{x}}) + \bm{z},  
\end{eqnarray}
where the matrix $\bm{H}$ is defined as 
\begin{eqnarray}
\bm{H}  = \bm{R} \frac{\partial \bm{g}}{\partial \bm{z}}.  
\end{eqnarray}




\section{Additional results}\label{sec:more_results}
In this section we provide additional results for both MNIST and CelebA dataset for different tasks discussed in the main paper.
\subsection{MNIST}
\begin{figure}[!ht]
\begin{center}
\subfigure[Digit 0]{\includegraphics[width=0.35 \linewidth]{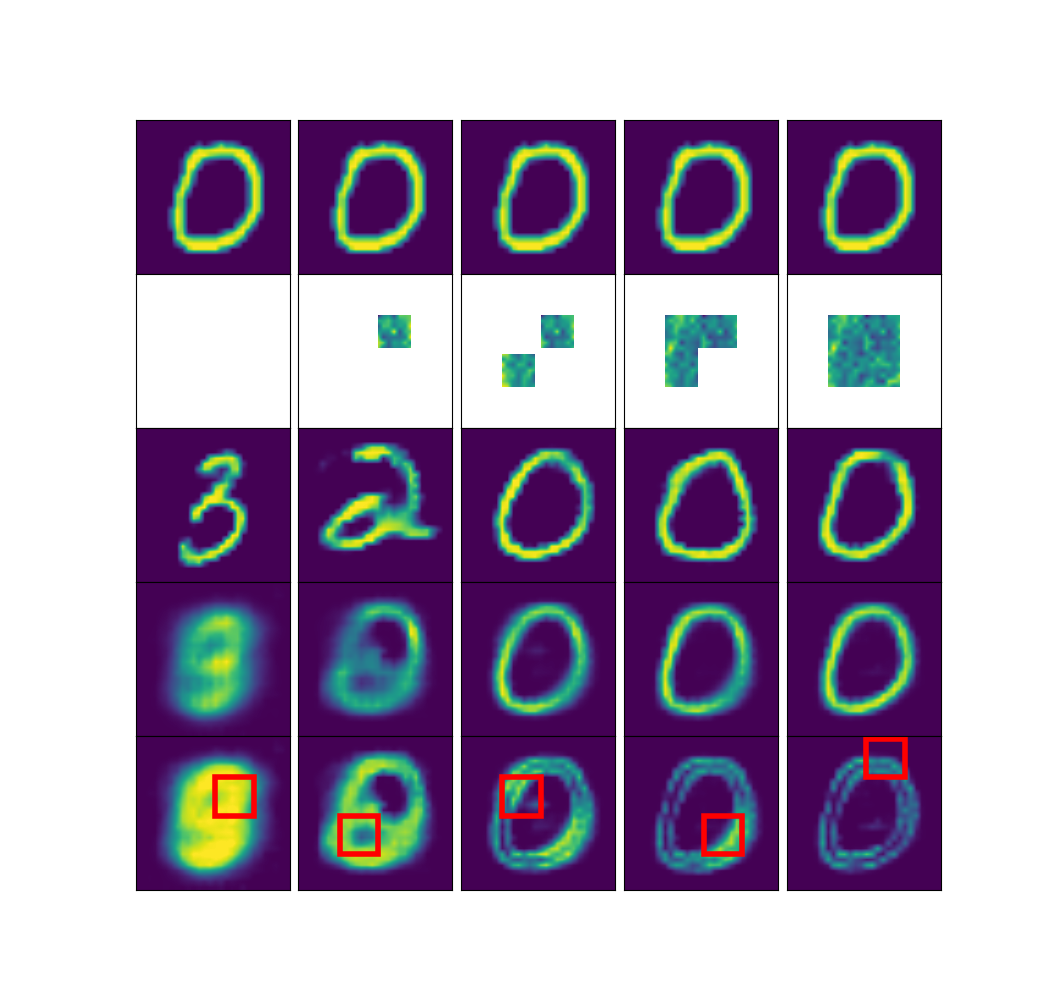}} 
\subfigure[Digit 1]{\includegraphics[width=0.35 \linewidth]{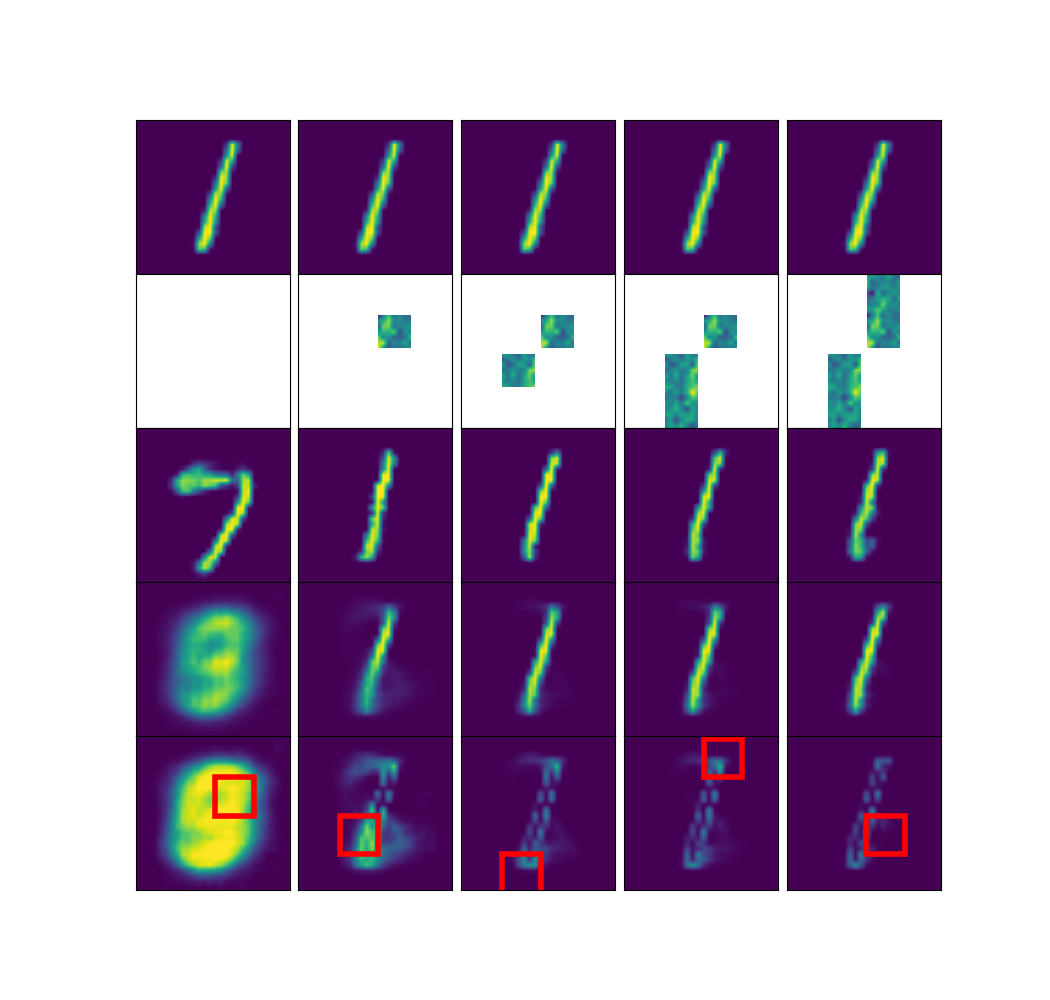}} 
\subfigure[Digit 2]{\includegraphics[width=0.35 \linewidth]{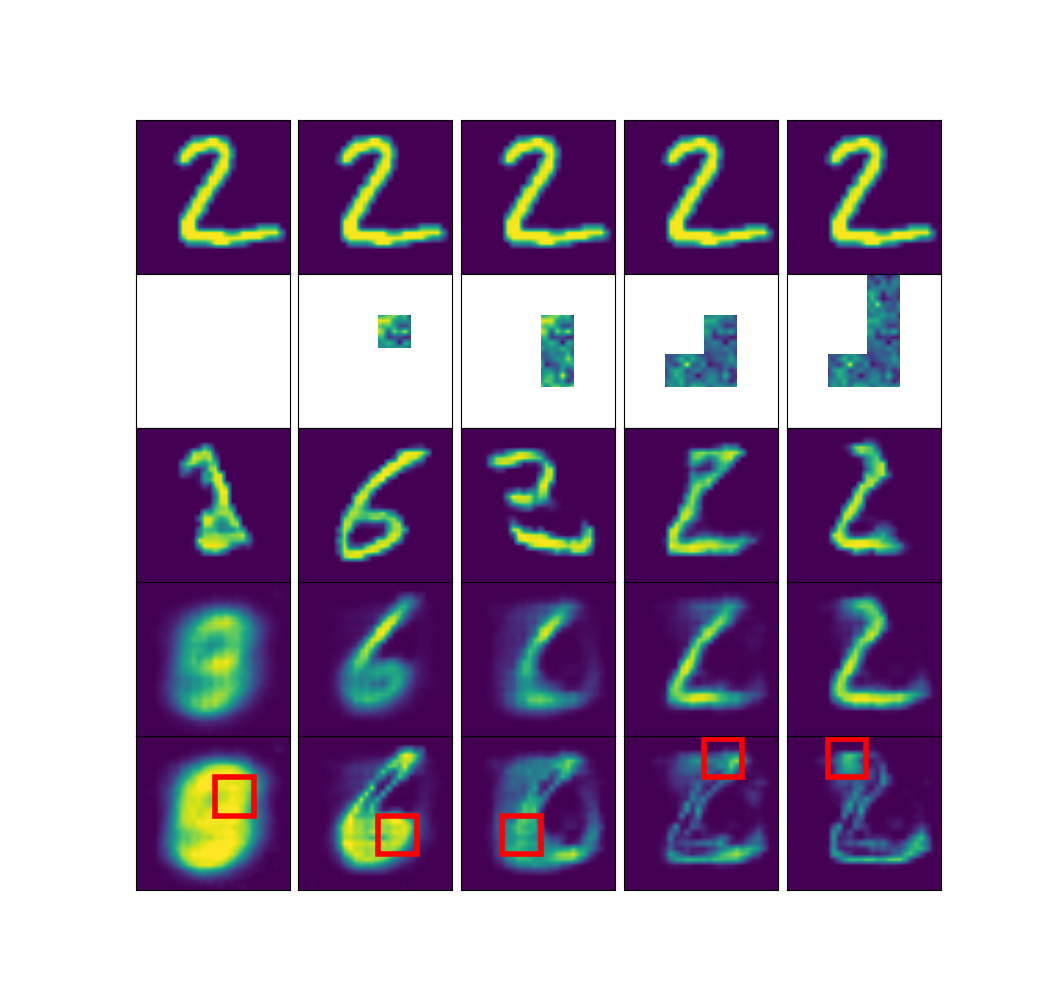}} 
\subfigure[Digit 3]{\includegraphics[width=0.35 \linewidth]{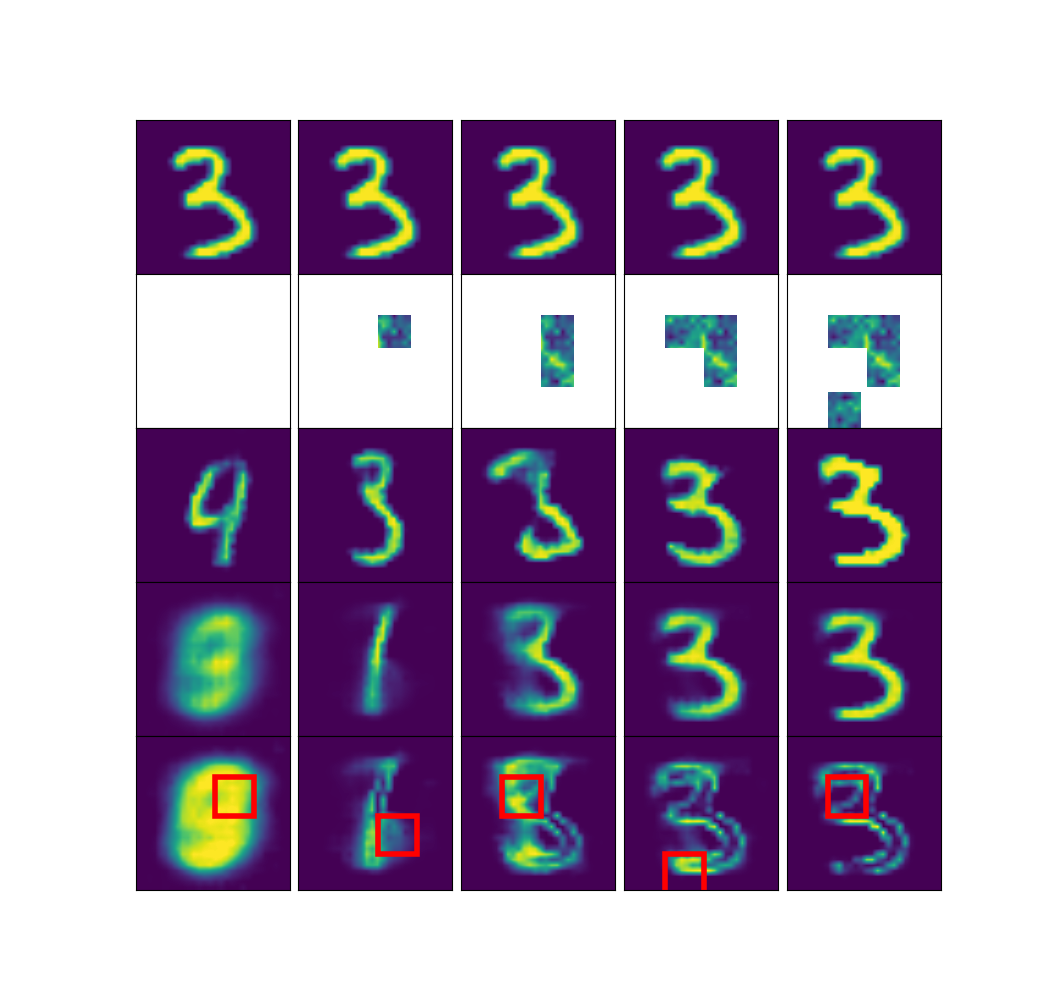}} 
\subfigure[Digit 4]{\includegraphics[width=0.35 \linewidth]{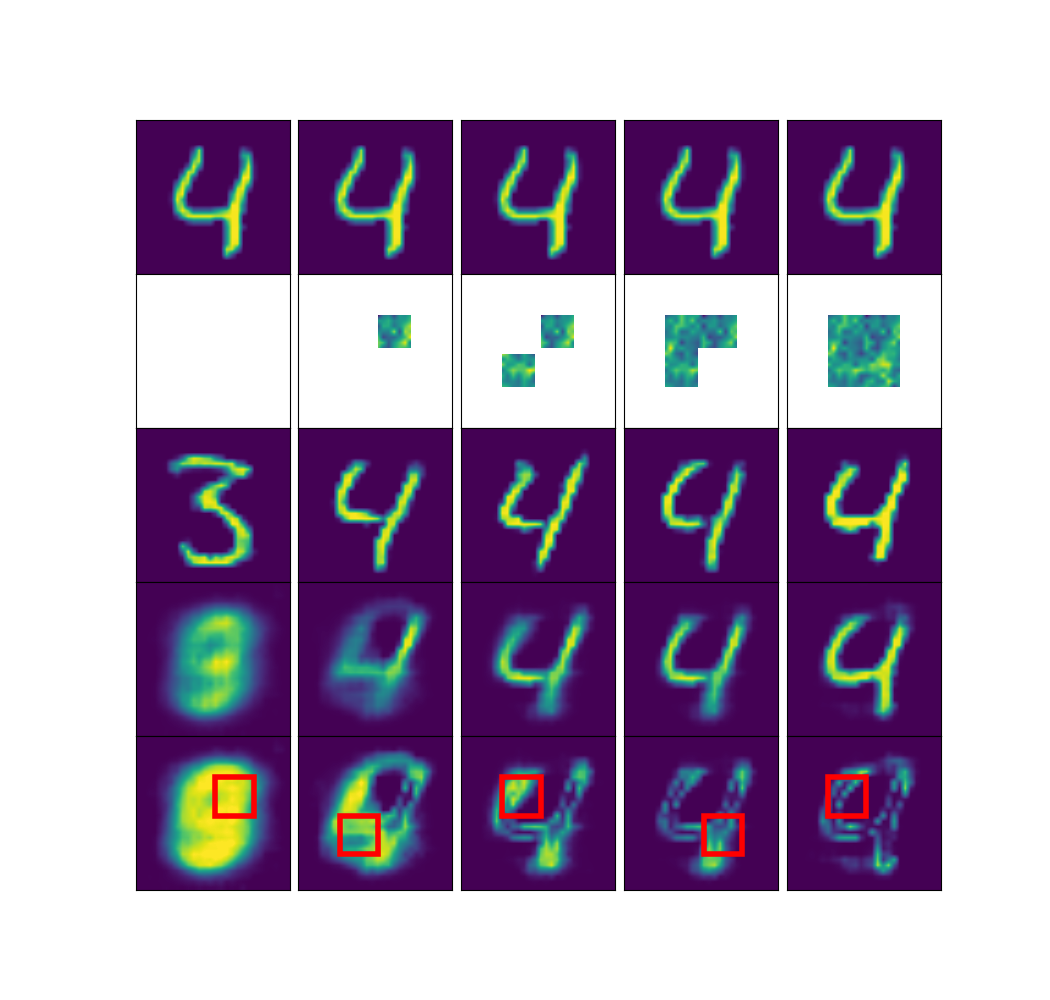}} 
\subfigure[Digit 6]{\includegraphics[width=0.35 \linewidth]{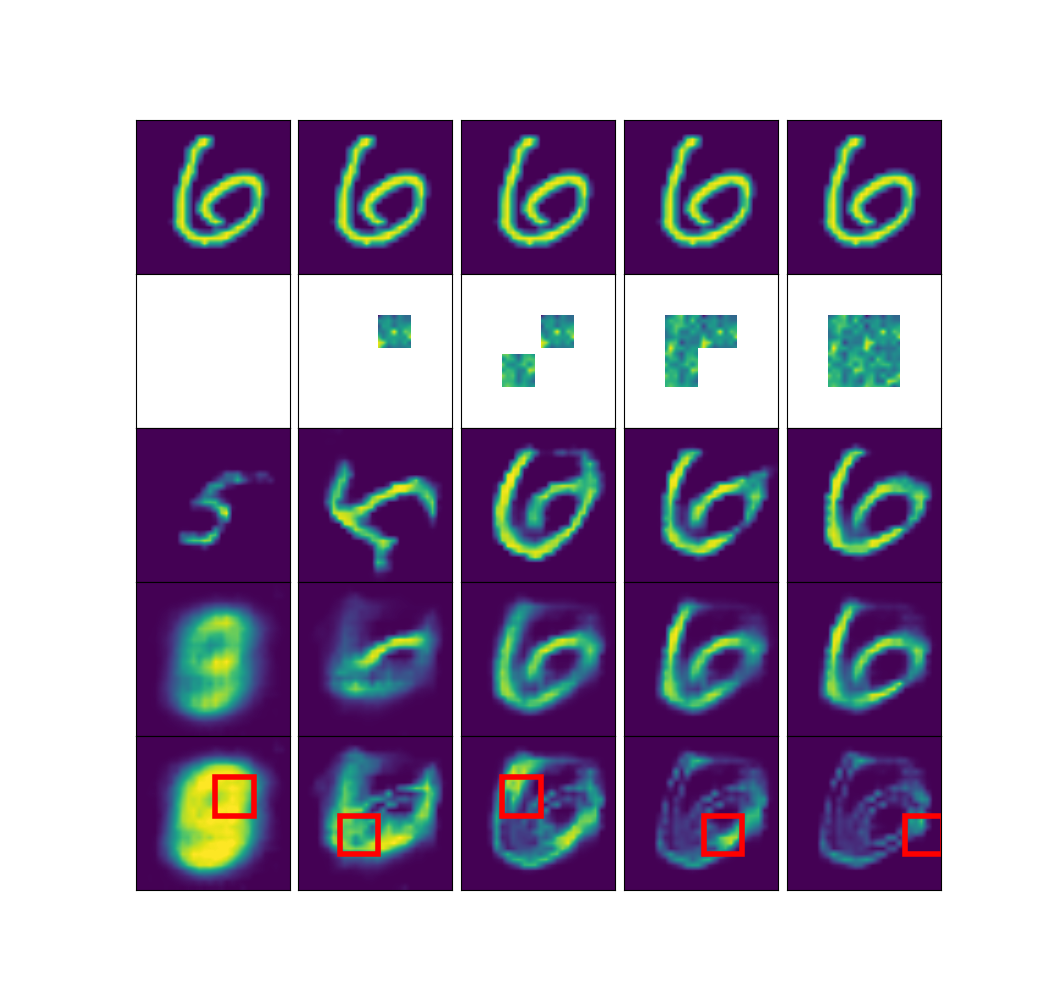}} 
\subfigure[Digit 7]{\includegraphics[width=0.35 \linewidth]{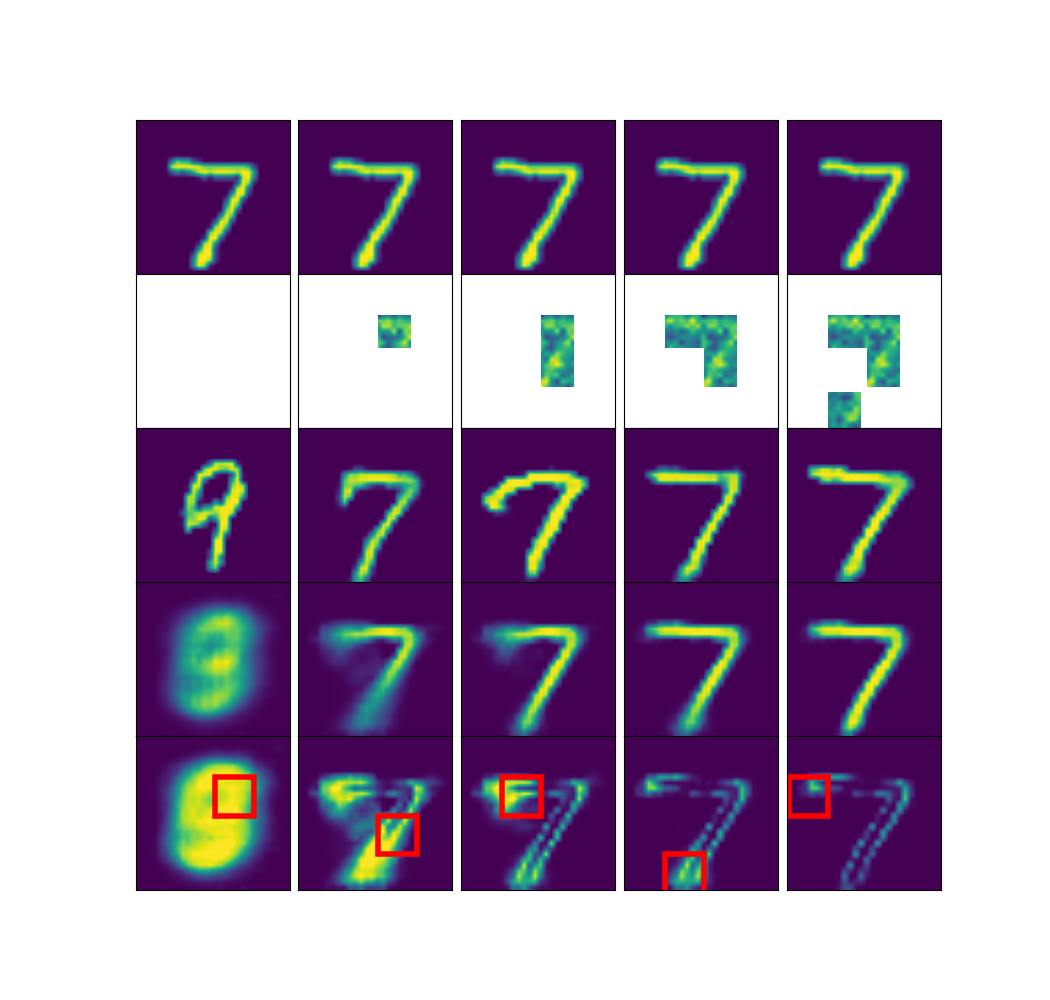}} 
\subfigure[Digit 9]{\includegraphics[width=0.35 \linewidth]{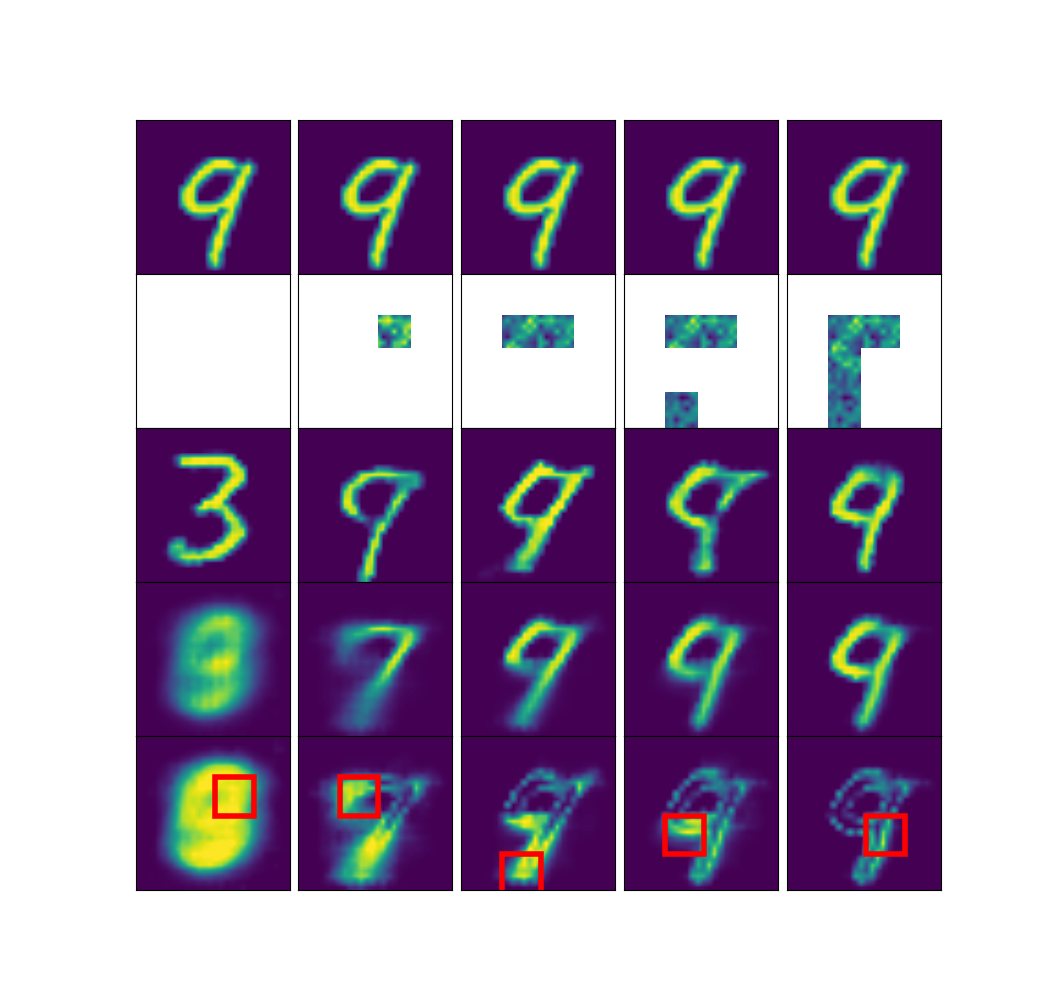}} 
\end{center}
\caption{\label{fig:oed_mnist} Estimate of the MAP (3rd row), mean (4th row) and variance (5th row) from the limited view of a noisy image (2nd row) using the proposed method. The window to be revealed at a given iteration (shown in red box) is selected using a variance-driven strategy. Top row indicates ground truth. For all digits $\sigma_y =1$.} 
\end{figure}

First we provide additional examples in Figure \ref{fig:oed_mnist} for variance-based adaptive measurement window selection procedure described in Section \ref{sec:img_recovery_mnist}.

\begin{figure}[!ht]
\begin{center}
\includegraphics[width=0.9 \linewidth]{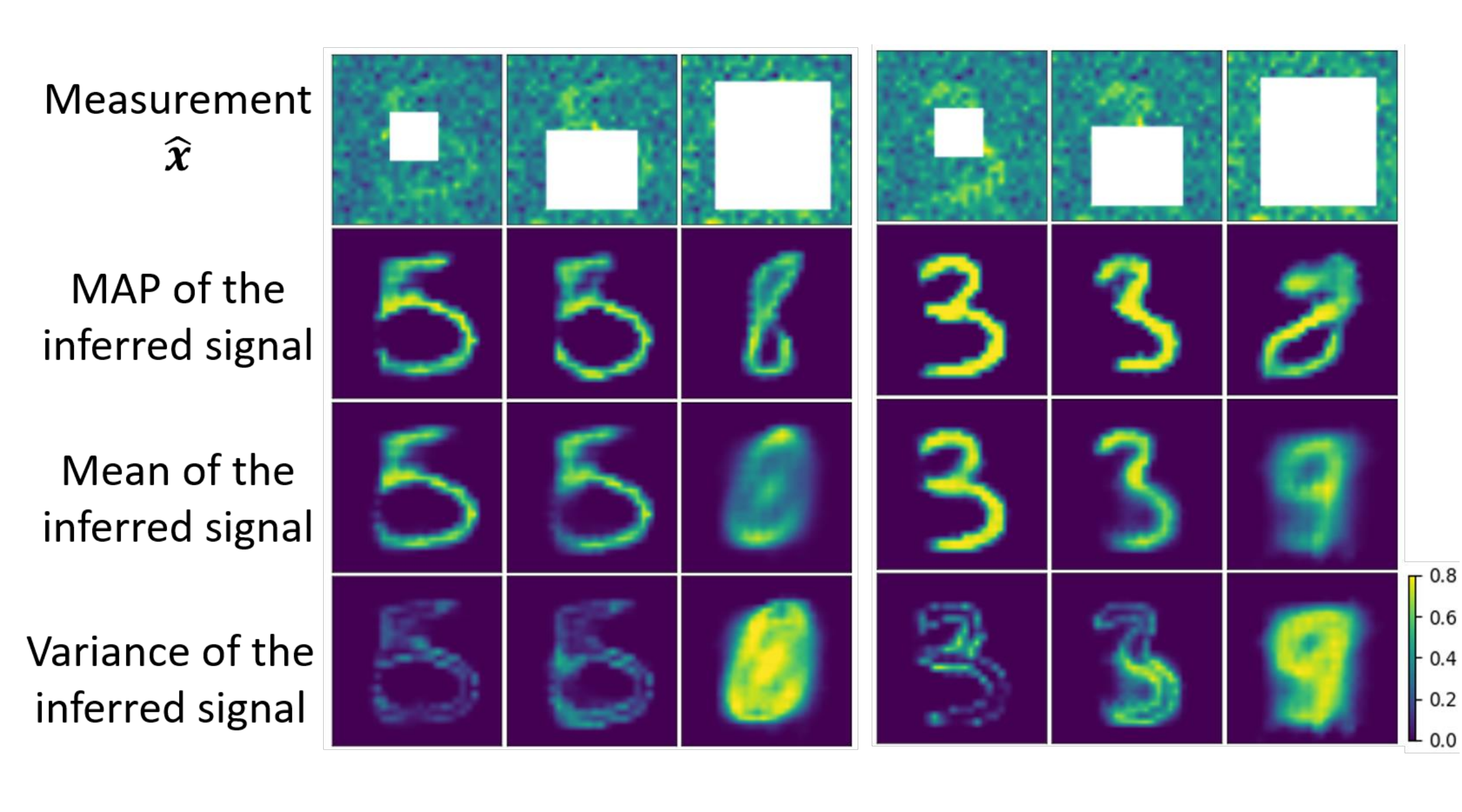} 
\end{center}
\caption{\label{fig:inpaint_mnist} Estimate of the MAP (2nd row), mean (3rd row) and variance (4th row) from a noisy image (1st row) using the proposed method. Note that all variance images are plotted on the same color scale and it highlights increasing level of uncertainty as more and more portion of an image is occluded.} 
\end{figure}

Figure \ref{fig:inpaint_mnist} shows additional results for the inpainting + denoising task, where an MNIST digit is occluded with masks of different sizes at different locations. Note that the variance is high where the occlusion mask is located indicating lower confidence in reconstructed image in that location. 

\subsection{CelebA} \label{ref:append:celeba}
For the CelebA dataset, we trained  WGAN-GP model using more than 200k celebrity facial images and perform inference using remaining test set images. The input images were cropped to a $64 \times 64$ RGB image and were normalized between [-1, 1]. 

\begin{figure}[!ht]
\begin{center}
\subfigure[]{\includegraphics[width=0.48 \linewidth]{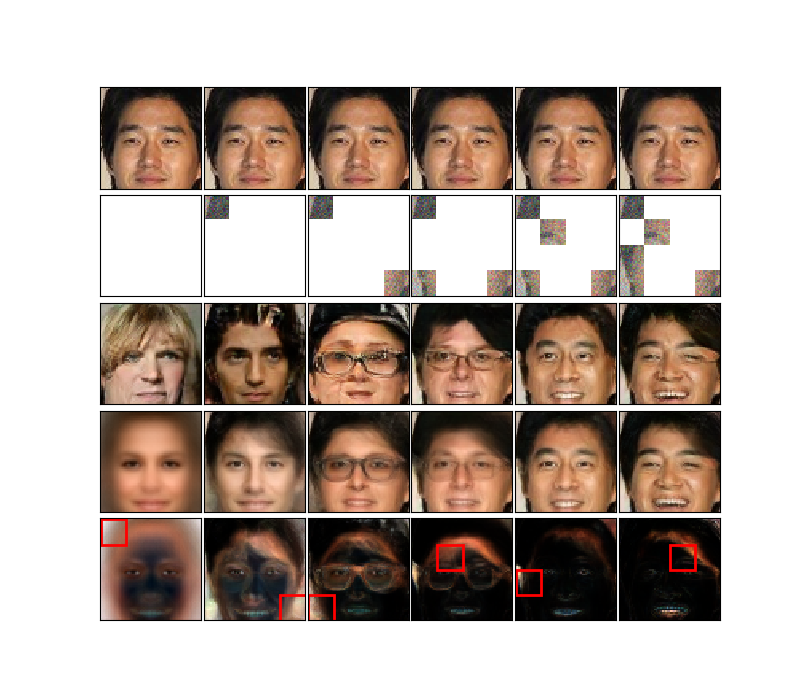}} 
\subfigure[]{\includegraphics[width=0.48 \linewidth]{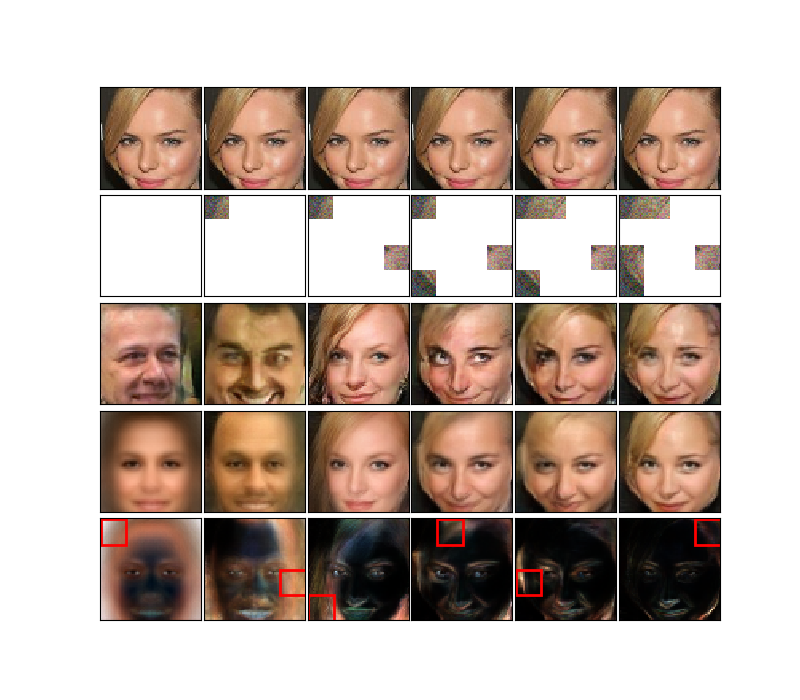}} 
\subfigure[]{\includegraphics[width=0.48 \linewidth]{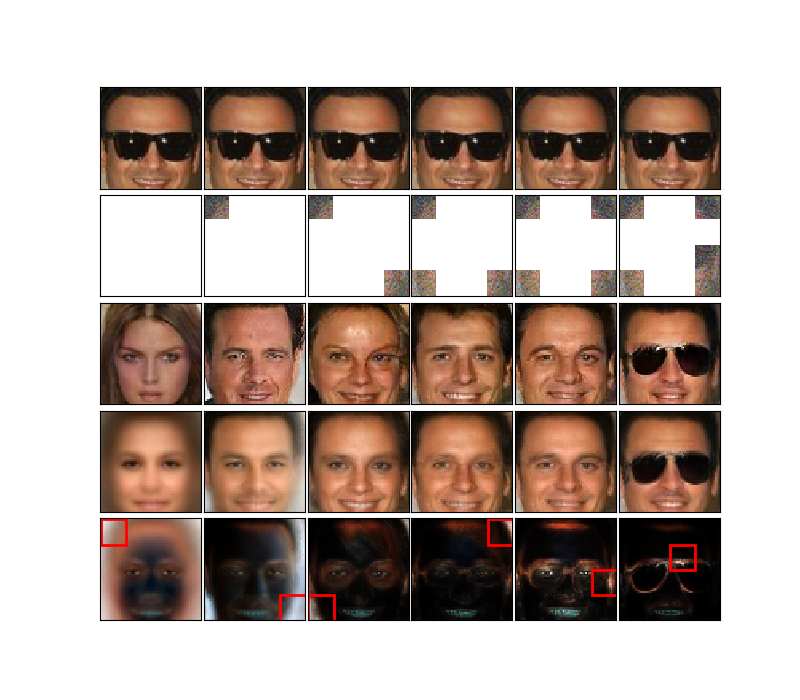}} 
\subfigure[]{\includegraphics[width=0.48 \linewidth]{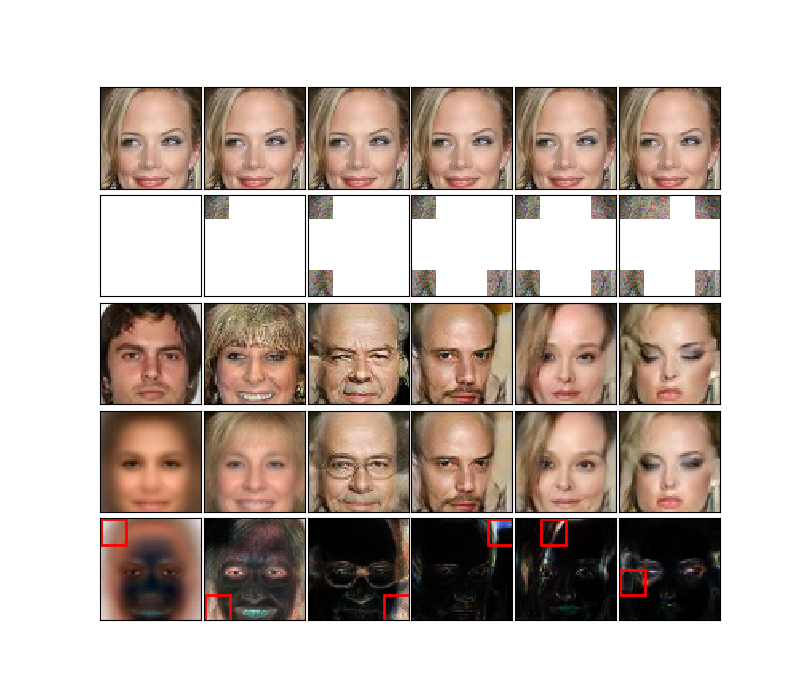}} 
\subfigure[]{\includegraphics[width=0.48 \linewidth]{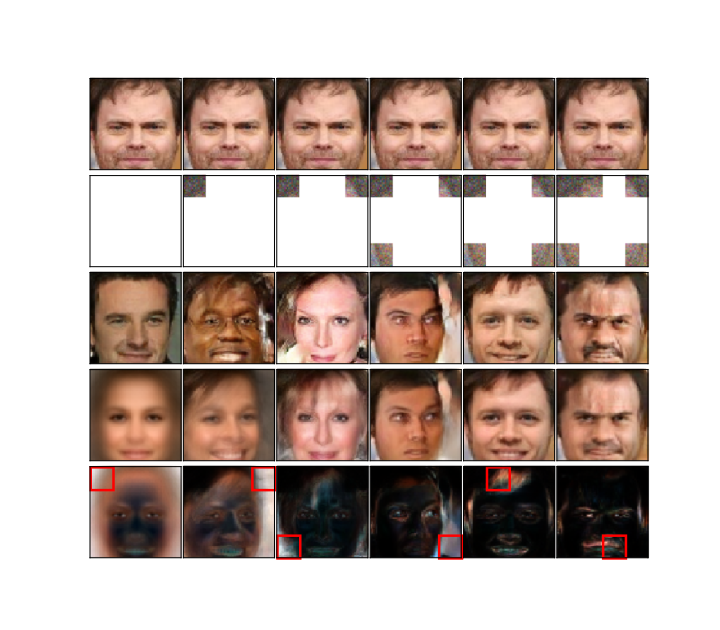}} 
\subfigure[]{\includegraphics[width=0.48 \linewidth]{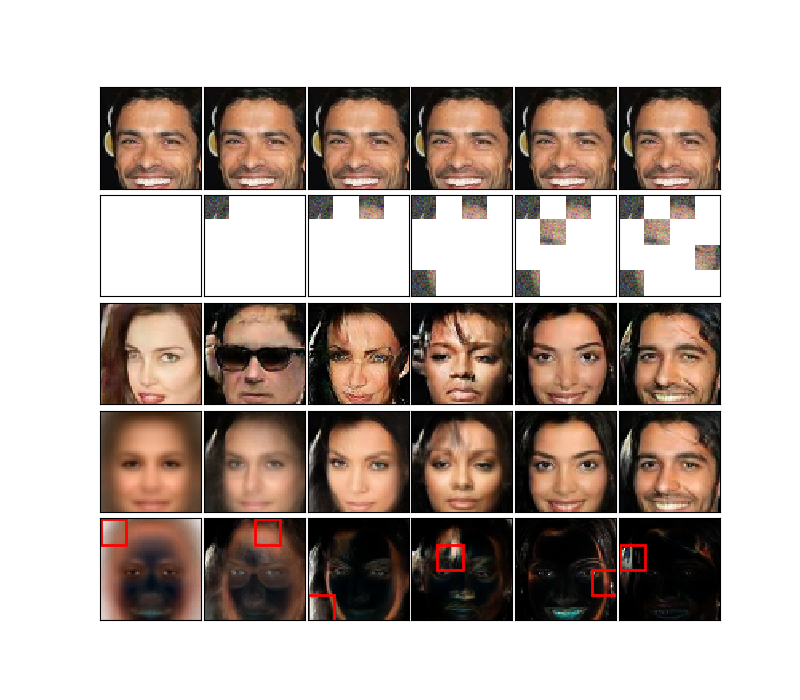}} 
\end{center}
\caption{\label{fig:oed_celeba} Estimate of the MAP (3rd row), mean (4th row) and variance (5th row) from the limited view of a noisy image (2nd row) using the proposed adaptive method. The window to be revealed at a given iteration (shown in red box) is selected using a variance-driven strategy. Top row indicates ground truth. For all images $\sigma_y = 1$.} 
\end{figure}

Once the GAN was trained, the HMC algorithm was used for posterior sampling and inference on a complimentary set of images (not used for training). In Figure \ref{fig:oed_celeba} we show some additional results for variance-based adaptive measurement window selection procedure for CelebA dataset. 

\begin{figure}[!ht]
\begin{center}
\subfigure[]{\includegraphics[width= \linewidth]{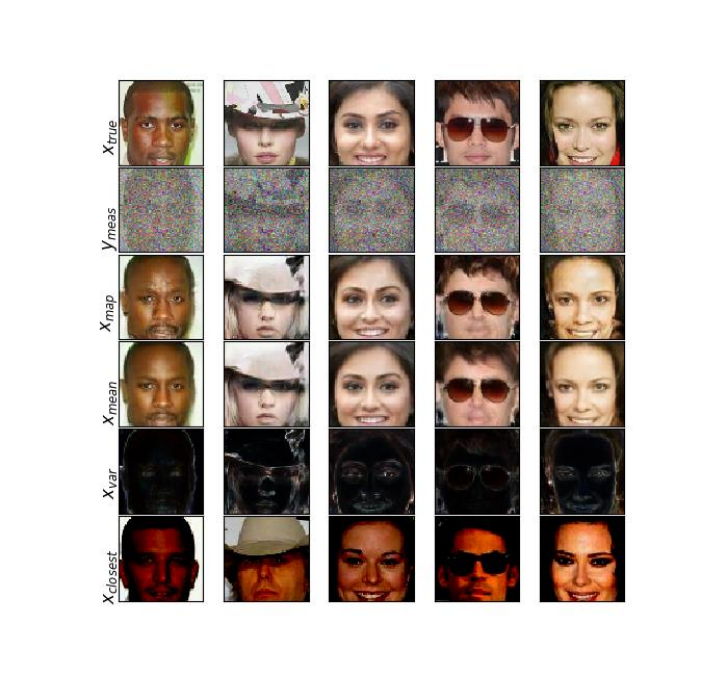}} 
\subfigure[]{\includegraphics[width= \linewidth]{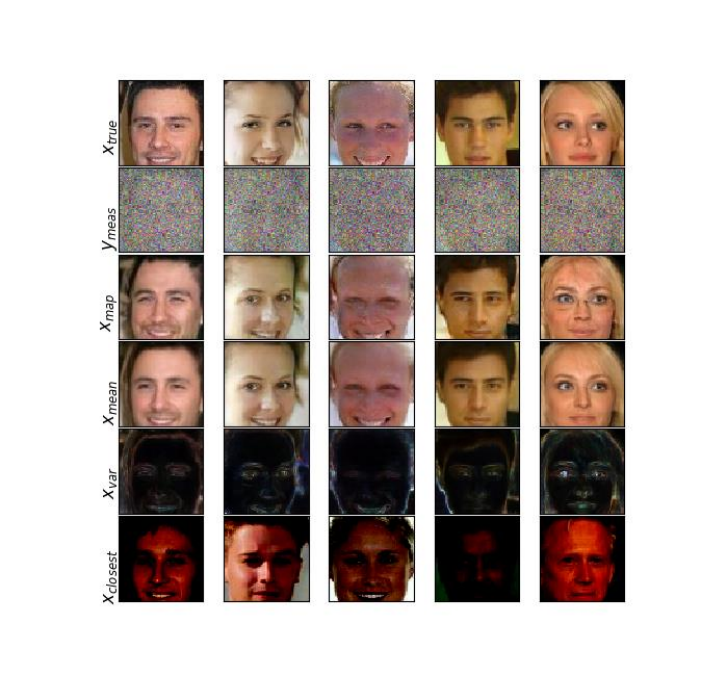}} 
\end{center}
\caption{\label{fig:recover_celeba} Estimate of the MAP (3rd row), mean (4th row) and variance (5th row) from a noisy image (2nd row) using the proposed method. Top row shows the ground truth. The last row shows the closest example in training set (by the $L_2$ measure). For all images $\sigma_y =1$.} 
\end{figure}

Next, in figure \ref{fig:recover_celeba} we show some additional results for image recovery task for CelebA dataset. Once again we note that the MAP estimate and the mean is close to the true image. On the other hand, the closest image from the training set (in an $L_2$ sense) is not as accurate. This points to the utility of using the GAN as an interpolant in the latent vector space.  

\section{Architecture and training details \label{sec:arch}}
We use the WGAN-GP model for learning prior density. The tuned value of hyper-parameters is shown in Table \ref{tab:hparams}.

\begin{table}[ht]
\renewcommand{\arraystretch}{1.5}
\centering
\caption{Hyper-parameters for WGAN-GP model}
\begin{adjustbox}{width=\linewidth}
\begin{tabular}{c c c c c c c}
\toprule
& \multicolumn{3}{c}{\textbf{CLASS 1 PROBLEMS}} & \multicolumn{3}{c}{\textbf{CLASS 2 PROBLEMS}} \\

\multirow{2}{*}{Task}  & \multicolumn{2}{c}{Image} & Physics-based & Image & \multicolumn{2}{c}{Image inpainting } \\
& \multicolumn{2}{c}{denoising} & inversion & classification & \multicolumn{2}{c}{and active learning} \\
Dataset & \multirow{2}{*}{MNIST}  & \multirow{2}{*}{CelebA}  & \multirow{2}{*}{synthetic}  & MNIST/ & \multirow{2}{*}{MNIST} & \multirow{2}{*}{CelebA}\\
& & & & NotMNIST & & \\

\midrule

Epochs & 1000 & 500 & 200 & 100 & 1000 & 500 \\
Learning rate & 0.0002 & 0.0001 & 0.0002 & 0.0002 & 0.0002 & 0.0001 \\
Batch size & 64 & 64 & 64 & 64 & 64 & 64 \\
$n_{critic}/n_{gen}$ & 5 & 5 & 1 & 2 & 5 & 5 \\
Momentum  & \multirow{2}{*}{0.5, 0.999} & \multirow{2}{*}{0.5, 0.999} & \multirow{2}{*}{0.5, 0.999} & \multirow{2}{*}{0.5, 0.999} & \multirow{2}{*}{0.5, 0.999} & \multirow{2}{*}{0.5, 0.999} \\
params. ($\beta_1$, $\beta_2$) & & & & & & \\

\bottomrule
\end{tabular}\label{tab:hparams}
\end{adjustbox}
\end{table}

\begin{figure}[h]
\begin{center}
\includegraphics[width=\linewidth]{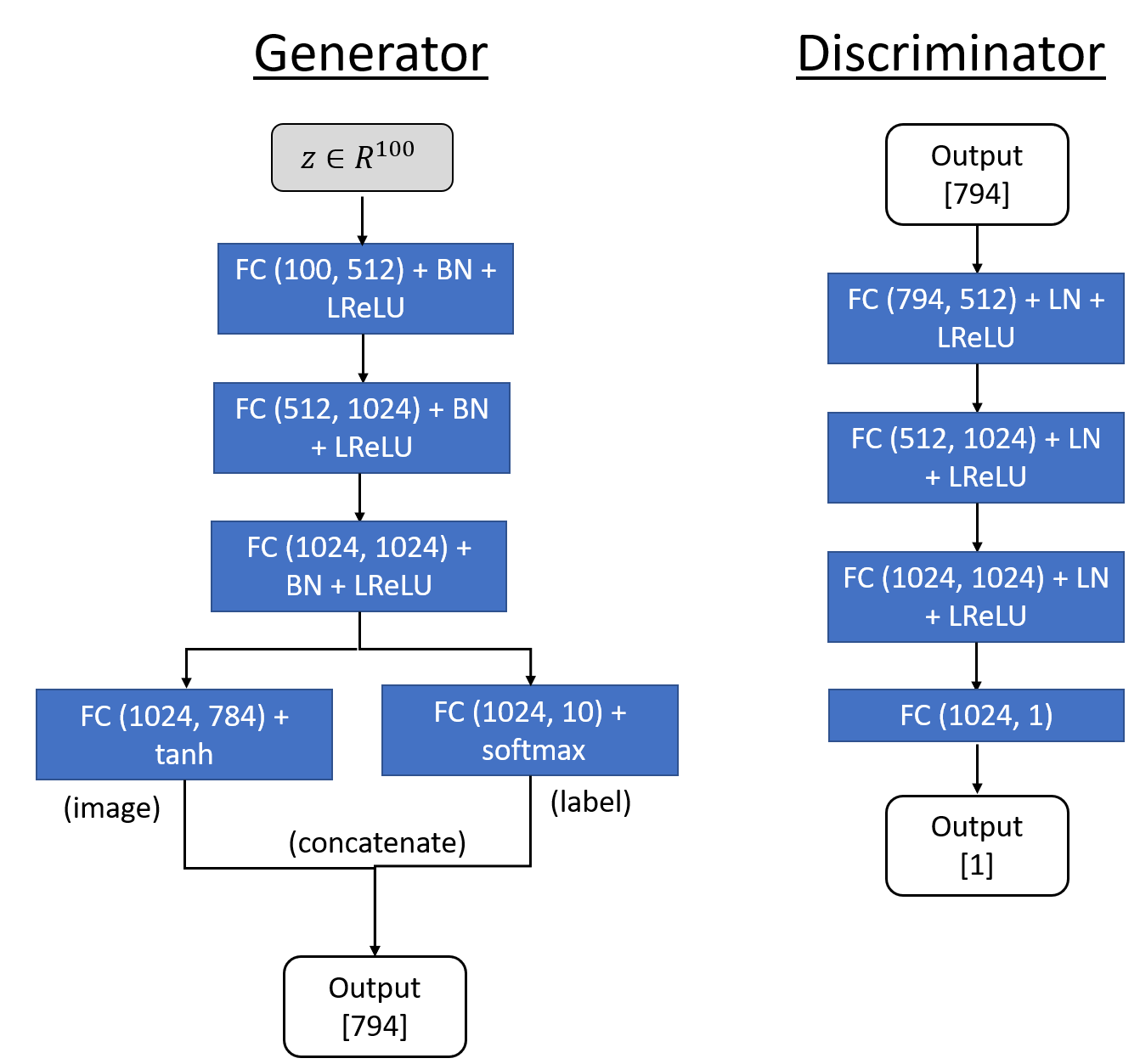} 
\end{center}
\caption{\label{fig:arch1} Generator and discriminator architecture used for image classification (hybrid modeling) task for both MNIST and NotMNIST datasaet.} 
\end{figure}

\begin{figure}[!ht]
\begin{center}
\subfigure[Architecture for MNIST and synthetic dataset (used in physics-based inference problem) ]{\includegraphics[width= \linewidth]{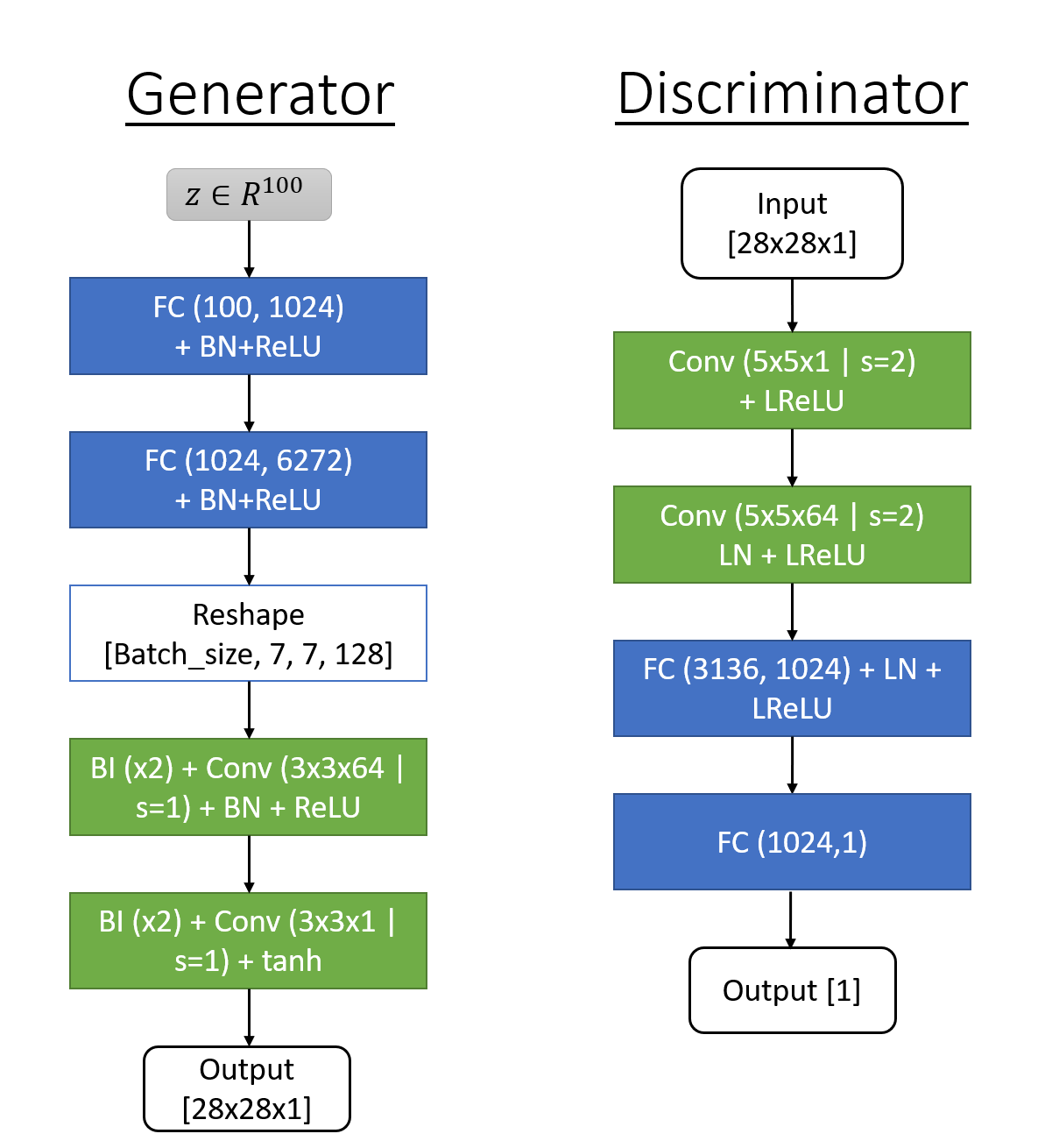}} 
\subfigure[Architecture for CelebA dataset]{\includegraphics[width= \linewidth]{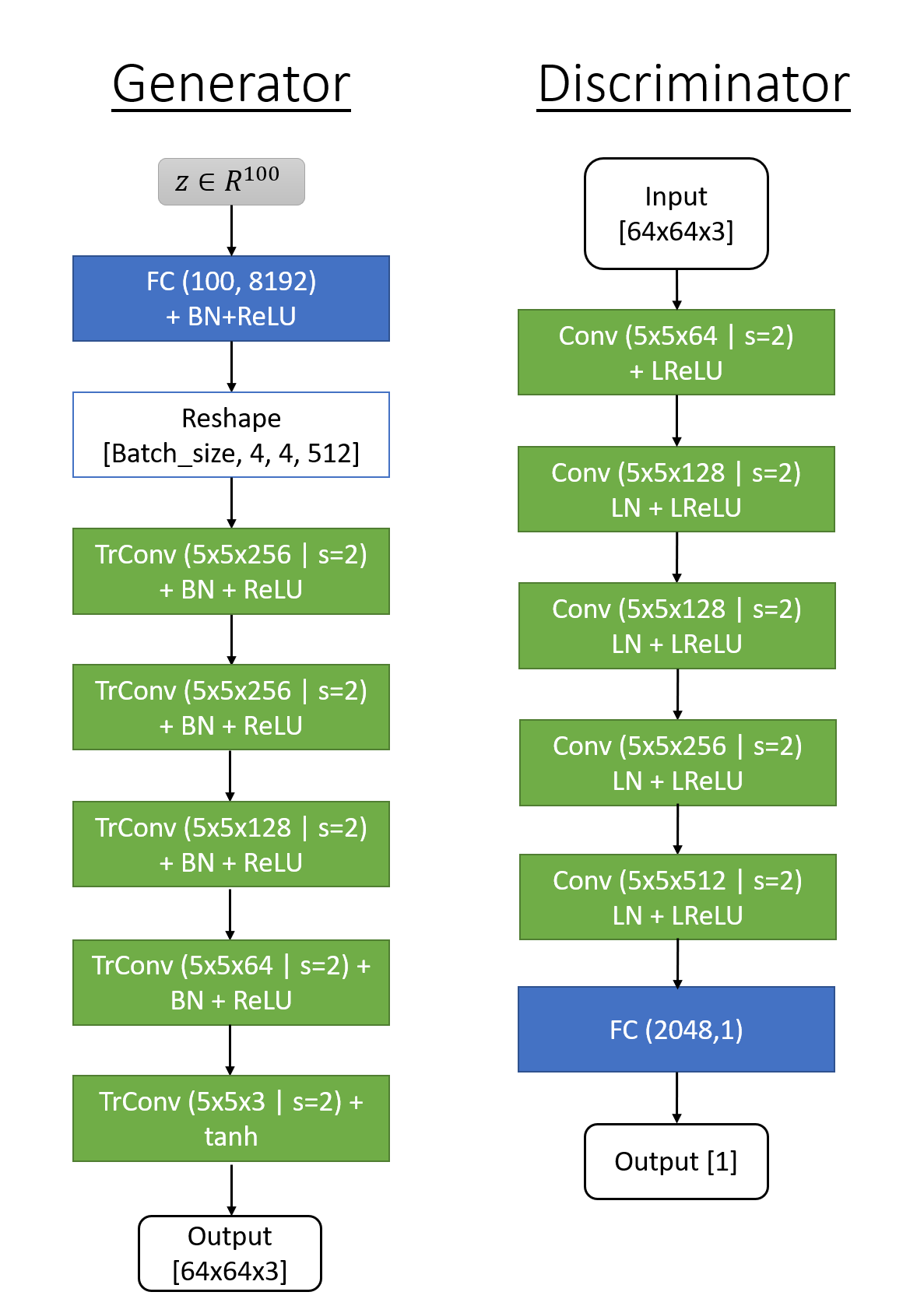}} 
\end{center}
\caption{\label{fig:arch2} Generator and discriminator architectures for (a) MNIST and synthetic dataset and (b) CelebA dataset used in image denoising, inapainting and physics-driven inversion.} 
\end{figure}

We use the same generator and discriminator architecture for the MNIST and the physics-based inference problem; whereas for the CelebA dataset we use a slightly different architecture to accommodate different input image size. The layout of both these architecture is shown in Figure \ref{fig:arch1} and \ref{fig:arch2}. Some notes regarding nomenclature used in  Figure \ref{fig:arch1} and \ref{fig:arch2}.

\begin{itemize}
    \item Conv (H $\times$ W $\times$ C $\vert$ s=n) indicates convolutional layer with filer size of HxW and number of filters=C with stride=n.
    \item FC (x,y) indicates fully connected layer with x neurons in input layer and y neurons in output layer.
    \item BN = Batch norm,  LN = Layer norm.
    \item TrConv = Transposed Convolution.
    \item LReLU = Leaky ReLU with $\alpha$=0.2.
\end{itemize}

\newpage

\end{document}